\documentclass[preprint,12pt]{elsarticle}

\usepackage{amssymb,latexsym,amsmath} 
\usepackage{bm}
\usepackage{graphicx} 
\usepackage{caption}
\usepackage{subcaption}
\usepackage{textcomp}
\usepackage{color}
\usepackage{todonotes}
\usepackage{blindtext}
\usepackage{ulem}

\journal{Journal of Computational Physics}

\title{A comparative study of physics-informed neural network models for learning unknown dynamics and constitutive relations}
\author{Ramakrishna Tipireddy, Paris Perdikaris, Panos Stinis and Alexandre Tartakovsky}
\address{Pacific Northwest National Laboratory, Richland WA 99354}
\date{today}

\begin{document}

\begin{abstract}
We investigate the use of discrete and continuous versions of physics-informed neural network methods for learning unknown dynamics or constitutive relations of a dynamical system . For the case of unknown dynamics, we represent all the dynamics with a deep neural network (DNN). When the dynamics of the system are known up to the specification of constitutive relations (that can depend on the state of the system), we represent these constitutive relations with a DNN. The discrete versions combine classical multistep discretization methods for dynamical systems with neural network based machine learning methods. On the other hand, the continuous versions utilize deep neural networks to minimize the residual function for the {\it continuous} governing equations. We use the case of a fedbatch bioreactor system to study the effectiveness of these approaches and discuss conditions for their applicability. Our results indicate that the accuracy of the trained neural network models is much higher for the cases where we only have to learn a constitutive relation instead of the whole dynamics. This finding corroborates the well-known fact from scientific computing that building as much structural information is available into an algorithm can enhance its efficiency and/or accuracy.  
\end{abstract}


\maketitle

\section{Introduction}
\label{sec:Introduction}

Due to the abundance of data and computational resources (e.g. graphical processing units), machine learning methods and, in particular, deep learning have gained prominence across various fields including scientific computing. In recent years there has been a lot of interest scientific computing community to utilize deep learning methods to accelerate scientific discovery. Combining the partial or complete knowledge about the physics of a system with the data through the use of deep learning has resulted in novel computational approaches \cite{PhysRevE.91.032915, chen2018, Han8505, SIRIGNANO20181339, felsberger2018, wan2018, MaE9994,raissi2019physics}.   

In the current work we employ both discrete and continuous models for learning unknown dynamics (physics) and constitutive relations of dynamical systems using physics informed neural networks (PINNs). Different versions of PINNs exist in the literature \cite{raissi2017physics1, raissi2017physics2, yang2018physics,tartakovsky2018learning}. We use a total of four different versions of PINNs to learn dynamics and constituive relations.

It is important to understand the applicability and the relative merits of these methods. In this work we aim to accomplish this task by testing these methods on the same physics model, namely a fedbatch bioreactor model (FBR) \cite{psichogios1992hybrid}. Although the methods discussed here are very general and can be used for any dynamical system modeled with ordinary differential equations (ODEs), we choose the FBR model, as it offers a dynamical system that is tractable but also very sensitive to initial conditions and variations of the problem parameters. 

The structure of this paper is as follows: In Section \ref{sec:methods} we discuss various discrete and continuous physics informed neural network (PINN) models for dynamical systems. We model a fedbatch (nonstationary) bioreactor (FBR) as a nonlinear dynamical system to test the various PINN models in Section \ref{sec:fedbatch}. In Section \ref{sec:results} we present numerical results for the various PINN models when applied to the FBR model.   Section \ref{sec:conclusions} contains conclusions and ideas for future work. 


\section{PINNs for dynamical systems}
\label{sec:methods}
We consider a dynamical system modeled as
\begin{equation} \label{eq:ode}
    \Dot{y}(t) = f(y(t),u(t),t;\lambda)
\end{equation}
where the state of the system at any time $t$ is given by the vector $y(t)\in \mathbb{R}^D$ and $f$ describes the dynamics of the system. Also, $u(t)$ describes possible external forcing and $\lambda$ is a potentially constitutive relation. We assume that the dynamics $f$ are at least partially unknown. Given the measurements of $y(t)$ at different time instants $t_1, t_2, \cdots, t_n$ we want to learn all the unknown parts of the dynamics $f$ using deep learning methods.  

\subsection{Neural network model} \label{sec:nn_model}
When the the dynamics of evolution ($f(y(t),u(t),t;\lambda)$) are completely unknown, we model the dynamics $f$ as a function of the states using a neural network (NN) and train the NN with time series data of the states. When the dynamics ($f(y(t),u(t),t;\lambda)$) of the system is known but the constitutive relation process ($\lambda(y(t))$) is not known we represent it using a NN.

\subsection{Multistep neural network model to learn unknown dynamics} \label{sec:multistepNN_rhs}
In this section we combine the multistep family of time stepping methods \cite{iserles2009first} for solving dynamical systems with PINNs \cite{raissi2017physics1, raissi2017physics2} to learn the unknown dynamics of a system from  time series data of the state. 

We consider a dynamical system of the form \eqref{eq:ode}. We discretize it in time using a generalized multistep method with $M$ steps \cite{raissi2018multistep} as follows
\begin{equation}\label{eq:multistep_dynamics}
    \sum_{m=0}^M [\alpha_m y_{n-m} + \Delta t \beta_m f(y_{n-m})] = 0, \; \; n=M, \cdots, N,
\end{equation}
where $y_{n-m}$ denotes the state of the system at time $t_{n-m}.$ Note that for a multistep method with $M$ steps we need to provide the first $M$ steps for the initialization of the method i.e. the values $y_0,\ldots,y_{M-1}.$  Different choices of $M$, $\alpha_m$ and $\beta_n$ lead to different schemes. For $M=1, \alpha_0=1, \alpha_1=-1$ and $\beta_0=\beta_1=0.5$ we find the trapezoidal rule given by
\begin{equation}\label{eq:multistep_loss}
    y_n  = y_{n-1} + \frac{1}{2} \Delta t [f(y_n) + f(y_{n-1})], \; \; n=1, \cdots, N. 
\end{equation}
If the dynamics of the system are completely unknown, then we approximate the function $f$ by a NN and learn the parameters of the neural network from the time series data. We do so by minimizing the mean square error loss function
\begin{equation}\label{eq:mse}
    MSE  = \frac{1}{N-M-1} \sum_{n=M}^N |l_n|^2, 
\end{equation}
where $l_n$ measures how well the NN approximation $f^{NN}$ of $f$ reproduces the exact (but unkonwn) dynamics of the ODE at time $t_n.$ The quantity  $l_n$ is given by 
\begin{equation}\label{eq:yn}
    l_n  = \sum_{m=0}^M [\alpha_m y_{n-m} + \Delta t \beta_m f^{NN}(y_{n-m})] , \; \; n=M, \cdots, N. 
\end{equation}
We train the NN representing  $f^{NN}$ so that the loss shown in \eqref{eq:mse} is minimized.

\subsection{Multistep neural network model to learn constitutive relations} \label{sec:multistepNN_mu}
In this section we employ a multistep neural network method to learn {\it only} the constitutive relation process ($\lambda(y(t))$) of the dynamical system instead of the whole dynamics ($f(y(t),u(t),t;\lambda$). This method is appropriate for situations where the governing equations of the dynamical system are known but certain constitutive relations of the system are unknown. In this case we represent the constitutive relation process ($\lambda(y(t))$) by a NN and train as before to minimize the mean square error loss $MSE.$

\subsection{Continuous model to learn unknown dynamics} \label{sec:contpinn_rhs}
In the continuous PINN model, when the dynamics $f$ is unknown, we model the state vector $y(t)$ as a function of time and the RHS describing the dynamics $f$ as a function of states $y(t)$ using NNs. We train these NN models using time series data of the states by minimizing the following loss function, 
\begin{gather}
    loss(f^{NN}(y^{NN}(t)))  = \frac{1}{N_y} \sum_{n=1}^{N_y} \left (y^{NN}(t_n)-y^*(t_n) \right )^2 \notag \\
    + \frac{1}{N_f} \sum_{n=1}^{N_f} \left (\frac{d y^{NN}}{dt}(t_n) - f^{NN}(y^{NN}(t_n)) \right )^2, \label{eq:loss_unknown_dynamics}
\end{gather}
Unlike the multistep neural network method the time series data doesn't have to be uniformly spaced. If we are interested in interpolation of the states, that is the solution within the time duration of the training data, we can use the NN $y^{NN}$. However, if we need to predict the solution beyond the time duration of the training data or with a different set of initial conditions, we need to use a time stepping scheme such as multistep or Runge-Kutta methods with the learned RHS ($f^{NN}(y(t))$ as the function describing the dynamics. The advantage of this approach is that we do not need the training data to be at equal intervals of time as is required for the multistep method. Another advantage is that we do not have to discretize the dynamical system for learning the constitutive relation or the dynamics.

\subsection{Continuous neural network model to learn constitutive relations} \label{sec:contPINN_mu}
In this section we employ a continuous physics-informed neural network method \cite{raissi2017physics1, raissi2017physics2} for learning constitutive relations of the system. In the discrete multistep neural network method the measurements of the state are required at uniform time intervals. This is necessitated by the multistep time stepping scheme. In the continuous PINNS model such condition can be relaxed and hence the measurements of the states can be obtained at arbitrary time intervals as is the case in many real experimental settings. For the dynamical system described in \eqref{eq:ode}, we represent the state of the system with a NN and minimize the following loss function to estimate the constitutive relation process $\lambda(y(t)),$
\begin{gather}
    loss(\lambda^{NN}(y^{NN}(t)))  = \frac{1}{N_y} \sum_{n=1}^{N_y} \left (y^{NN}(t_n)-y^*(t_n) \right )^2 \notag \\ 
    + \frac{1}{N_f} \sum_{n=1}^{N_f} \left (\frac{d y^{NN}}{dt}(t_n) - f(y^{NN}(t_n),\lambda^{NN}(y^{NN}(t_n))) \right )^2, \label{eq:loss_unknown_parameters}
\end{gather}
where $N_y$ is the number of measurements of the system state and $N_f$ is the total number of pre-determined collocation points.

\section{Fedbatch bioreactor model} \label{sec:fedbatch}
Dynamical systems describing biological phenomena are very complex and can be of the form described in \eqref{eq:ode}
where the state vector $y$ and control vector $u$ depend on the constitutive relation $\lambda$ which in turn depends on state and control vectors as follows
\begin{equation} \label{eq:dynpara}
	\lambda = g(y,u), 
\end{equation}
for some function $g.$
Such an interdependence is more often than not expressed through  nonlinear terms and the resulting system is often chaotic. The dependence of the constitutive relation $\lambda$ on the state and control vectors is in general unknown and is difficult to derive from first principles due to complex biological reaction kinetics. Hence one has to estimate these constitutive relations by indirect means using experimental data. Inaccurate modeling of constitutive relations results in inaccurate predictions of the state vector. 

Here we consider a bioreactor operating in Fedbatch (nonstationary) conditions in which the microbial growth exhibits a wide range of dynamical behavior \cite{psichogios1992hybrid} due to the continuous change of the growth rate in a complex manner. The Fedbatch bioreactor (FBR) model is given by
\begin{align}
  \frac{d X(t)}{\partial t} &= \mu(X(t),S(t),V(t)) X(t)) - \frac{F(t) X(t)}{V(t)}, \notag \\  
  \frac{d S(t)}{\partial t} &=-k_1 \mu(X(t),S(t),V(t)) X(t))X(t) + \frac{F(t)(S_{in}(t)-S(t))}{V(t)}, \label{eq:fedbatch} \\
  \frac{d V(t)}{\partial t} &= F(t), \notag
\end{align}
subject to the intitial conditions
\begin{equation}
X(0)=X_0, \quad S(0)=S_0,\quad V(0)=V_0,   
\end{equation}
where $X(t)$ is the biomass concentration, $S(t)$ is the substrate concentration, and $V(t)$ is the volume of the bioreactor. The dynamics of the bioreactor are described by the mass balance between the reacting species and the kinetics of the specific {\it growth rate} $\mu(X(t),S(t),V(t)) X(t))$ which accounts for the rate at which the substrate is converted to the biomass. Other parameters in the governing equations are the inlet substrate concentration $S_{in}(t)$, the flow rate $F(t)$ and the substrate to cell conversion coefficient $k_1$. Although it is very difficult to model the specific growth rate $\mu(X(t),S(t),V(t)) X(t))$ due to its dependence on the states of the bioreactor system, we consider the Haldane model from \cite{psichogios1992hybrid} as the ground truth model:
\begin{equation} \label{eq:haldane}
    \mu(S(t)) = \frac{\mu^* S(t)}{K_m + S(t) + \frac{S(t)^2}{K_i}}
\end{equation}
where $K_m$ and $K_i$ are model constants. In the numerical experiments we consider the Haldane model \eqref{eq:haldane} for $\mu(t)$ and solve the system \eqref{eq:fedbatch} to synthetically generate the data required to train the neural network. 


\subsection{FBR model with mulitstep NN for learning the unknown dynamics $f$} \label{sec:fbr_multistep_f}
Let the state vector of the dynamical system be $y(t) = [X(t), S(t), V(t)]^T$ and the RHS of the dynamical system is described by the unknown function $f(y,t) = [f_1(y,t), f_2(y,t), f_3(y,t)]^T$ such that 
\begin{align}\label{eq:multistepNN_rhs}
  \frac{\partial X(t)}{\partial t} &= f_1[X(t), S(t), V(t),t], \nonumber \\  
  \frac{\partial S(t)}{\partial t} &=f_2[X(t), S(t), V(t),t], \nonumber \\
\frac{\partial V(t)}{\partial t} &= f_3[X(t), S(t), V(t),t].
\end{align}
Using vector notation we have
\begin{equation} \label{eq:multistepNN_y}
    \frac{d y}{d t} = f(y(t))
\end{equation}
%
%
Here the RHS function $f(y(t))$ is modeled as a neural network $f^{NN}(y(t))$ that takes the state $y(t)$ as input and outputs the NN approximation $\hat{f}(y(t))$ to the true RHS. The NN-produced function $\hat{f}(y(t))$ is used to solve the ODE using multistep method to obtain the solution of the learned dynamical system as follows
\begin{equation}\label{eq:multistepNN}
    \sum_{m=0}^M [\alpha_m y_{n-m} + \Delta t \beta_m f^{NN}(y_{n-m})] = 0,\; \;  n=M, \ldots, N.
\end{equation}
%

\subsection{FBR model with mulitstep NN for learning the constitutive relation $\mu(t)$}
Here the constitutive relation $\mu(t)$ is modeled as a neural network ($\mu^{NN}(y)$) that takes the state $y(t)$ as input and outputs the value of the constitutive relation. The NN-estimated $\mu^{NN}(y)$ is substituted into the governing differential equations and we find
\begin{align}
  \frac{d X(t)}{\partial t} &= \mu^{NN}(t) X(t) - \frac{F(t) X(t)}{V(t)}, \notag \\  
  \frac{d S(t)}{\partial t} &=-k_1 \mu^{NN}(y(t)) X(t) + \frac{F(t)(S_{in}(t)-S(t))}{V(t)}, \label{eq:fedbatch_parameter} \\
  \frac{d V(t)}{\partial t} &= F(t), \notag
\end{align}
The system \eqref{eq:fedbatch_parameter} was solved using the multistep method described in Section \ref{sec:multistepNN_rhs}. Note that the neural network $\mu^{NN}(y(t))$ takes {\it all the components} of the state vector $y(t) = [X(t),S(t),V(t)]^T$ as input and returns the prediction $\mu^{NN}(y(t)).$  If we have further knowledge that this constitutive relation $\mu$ depends {\it only} on the  state $S(t)$ as e.g. in the Haldane model \eqref{eq:haldane}, we can design a smaller neural network $\mu^{NN}(S(t))$ which can potentially require a smaller amount of data to train without affecting the accuracy of the prediction. However, for the sake of generality we assume that such knowledge about $\mu$ is not available and design a neural network that depends on all the components of the state vector. 

\subsection{FBR model with continuous PINNs for learning dynamics $f$} \label{sec:contpinn_rhs_fbr}
In the continuous PINN model, when the dynamics $f$ is unknown, we model the state vector $y(t)$ as a function of time and the RHS describing the dynamics $f$ as a function of states $y(t)$ using NNs. We train these NN models using time series data of the states by minimizing the following loss function, 
\begin{gather}
    loss(f^{NN}(y^{NN}(t)))  = \frac{1}{N_y} \sum_{n=1}^{N_y} \left (y^{NN}(t_n)-y^*(t_n) \right )^2 \notag \\
    + \frac{1}{N_f} \sum_{n=1}^{N_f} \left (\frac{d y^{NN}}{dt}(t_n) - f^{NN}(y^{NN}(t_n)) \right )^2, \label{eq:loss_unknown_dynamics_fbr}
\end{gather}
Unlike the multistep neural network method the time series data doesn't have to be uniformly spaced. If we are interested in interpolation of the states, that is the solution within the time duration of the training data, we can use the NN $y^{NN}$. However, if we need to predict the solution beyond the time duration of the training data or with a different set of initial conditions, we need to use a time stepping scheme such as multistep or Runge-Kutta methods with the learned RHS ($f^{NN}(y(t))$ as the function describing the dynamics.   

\subsection{FBR model with continuous PINNs for learning the constitutive relation $\mu(t)$} \label{sec:contpinn_mu}
In the current formulation of continuous PINN model we model the state vector $y(t)$ (notation from \ref{sec:fbr_multistep_f}) and the constitutive relation process $\mu(t)$ using neural networks and minimize the following loss function 
\begin{gather}
    loss(\mu^{NN}(y^{NN}(t)))  = \frac{1}{N_y} \sum_{n=1}^{N_y} \left (y^{NN}(t_n)-y^*(t_n) \right )^2 \notag \\
    + \frac{1}{N_f} \sum_{n=1}^{N_f} \left (\frac{d y^{NN}}{dt}(t_n) - f(y^{NN}(t_n),\mu^{NN}(y^{NN}(t_n))) \right )^2, \label{eq:loss_unknown_parameters_fbr}
\end{gather}
such that $\mu^{NN}(y^{NN}(t_n))$ estimates the constitutive relation as a function of the state and $y^{NN}(t)$ predicts the state vector as a function of time $t$.  Here, we can use the NN model $y^{NN}(t)$ to predict the solution (interpolate) within the time duration of the training data. However, to find the solution for different time duration and different initial conditions, one has to solve the governing equations \eqref{eq:fedbatch_parameter} using the NN model $\mu^{NN}(y).$

\section{Numerical Experiments}
\label{sec:results}
We present numerical results for the FBR model with constitutive relations $k_1=1,K_M=10,K_i=0.1$ and $\mu^*=5.$ \cite{dochain1990adaptive}. The initial conditions used for generating training data as a solution of the governing equations \eqref{eq:fedbatch} are $X_0 = 0.1,$ $S_0 = 1,$ and $V_0 = 10$ \cite{dochain1990adaptive}. Without loss of generality, we use $F(t) = 0.1$ and $S_{in}(t) = 3.5.$ For learning the dynamics $f(y)$ and constitutive relation  $\mu(y)$, we use the data synthesized using the Haldane model \eqref{eq:haldane} to train the neural networks. 

\subsection{Discrete time - Unknown dynamics}

\begin{figure}[h!]
    \centering
    \begin{subfigure}[t]{0.48\textwidth}
        \centering
        \includegraphics[scale=.5]{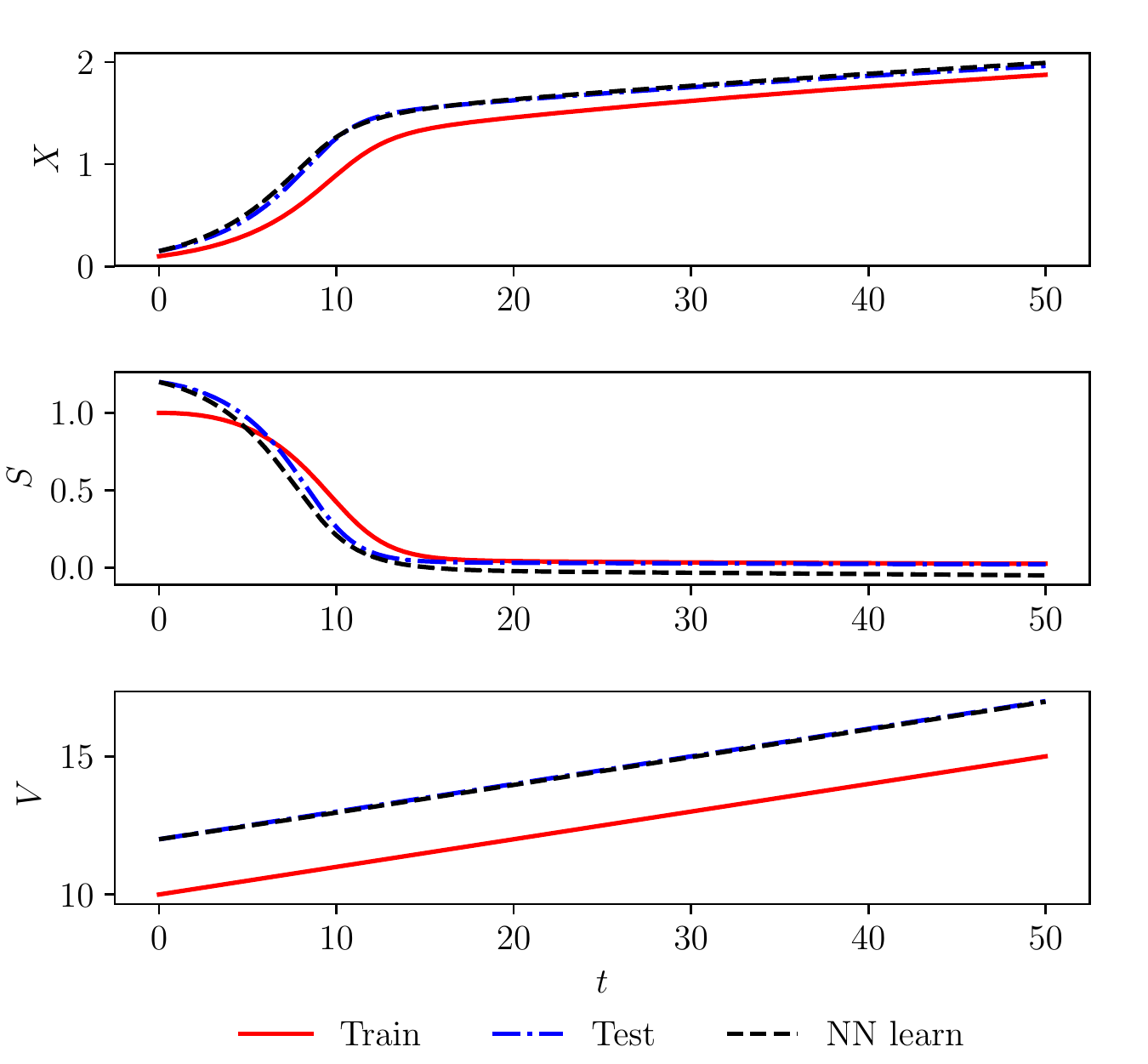}
        \caption{} \label{fig:multistepnn_rhs_state}
    \end{subfigure}        
    \begin{subfigure}[t]{0.48\textwidth}
        \centering
        \includegraphics[scale=.5]{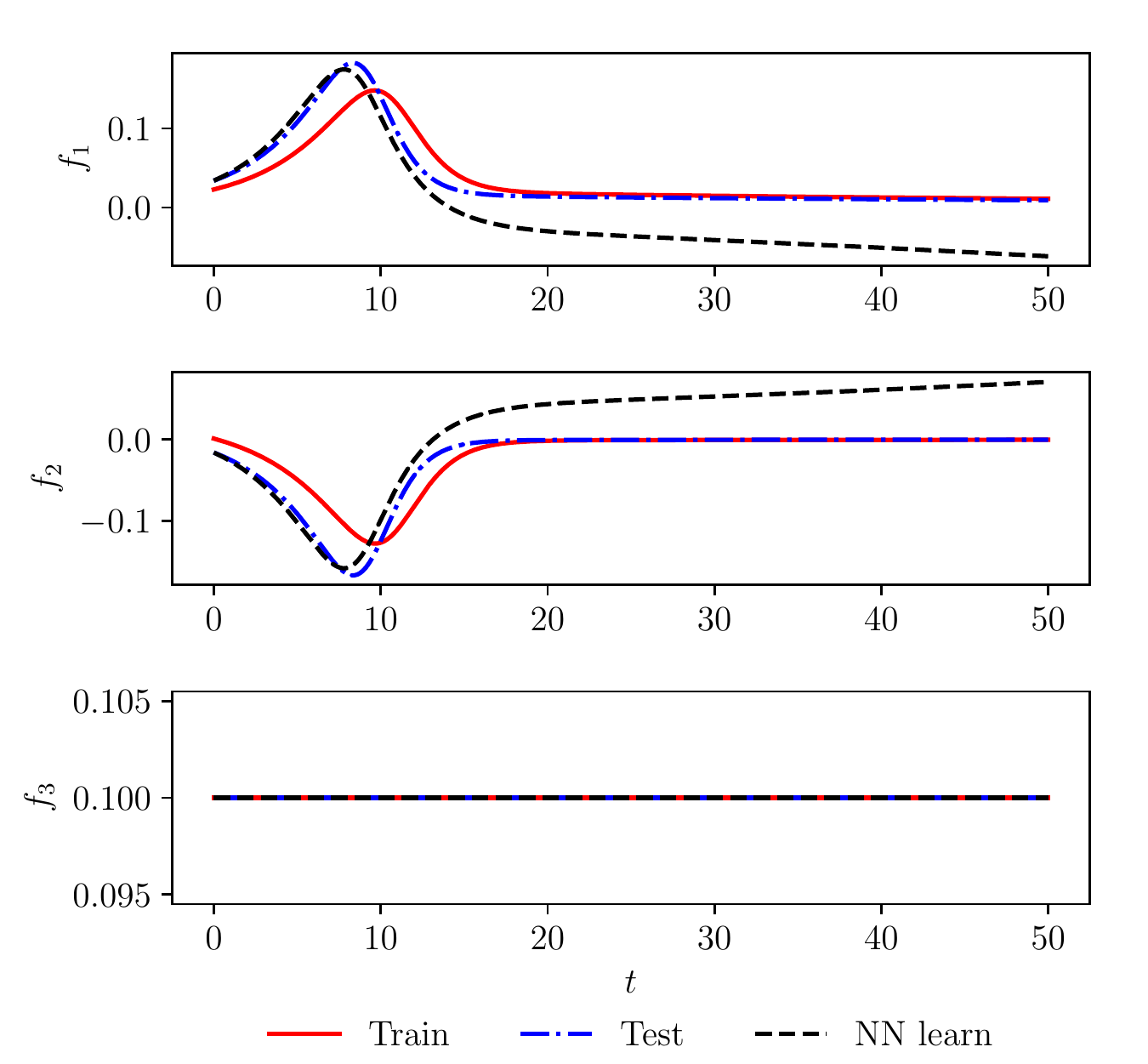}
        \caption{} \label{fig:multistepnn_rhs_rhs}
    \end{subfigure}    
      \caption{Unknown dynamics - Multistep NN (1 trajectory used for training): a). Solution of FBR model computed from exact dynamics and with learned dynamics using multistep neural networks, b). Right hand side (RHS) of the ODE describing the evolution dynamics.}   \label{multistepnn_rhs}
\end{figure}

\begin{figure}[h!]
    \centering
    \begin{subfigure}[t]{0.48\textwidth}
        \centering
        \includegraphics[scale=.42]{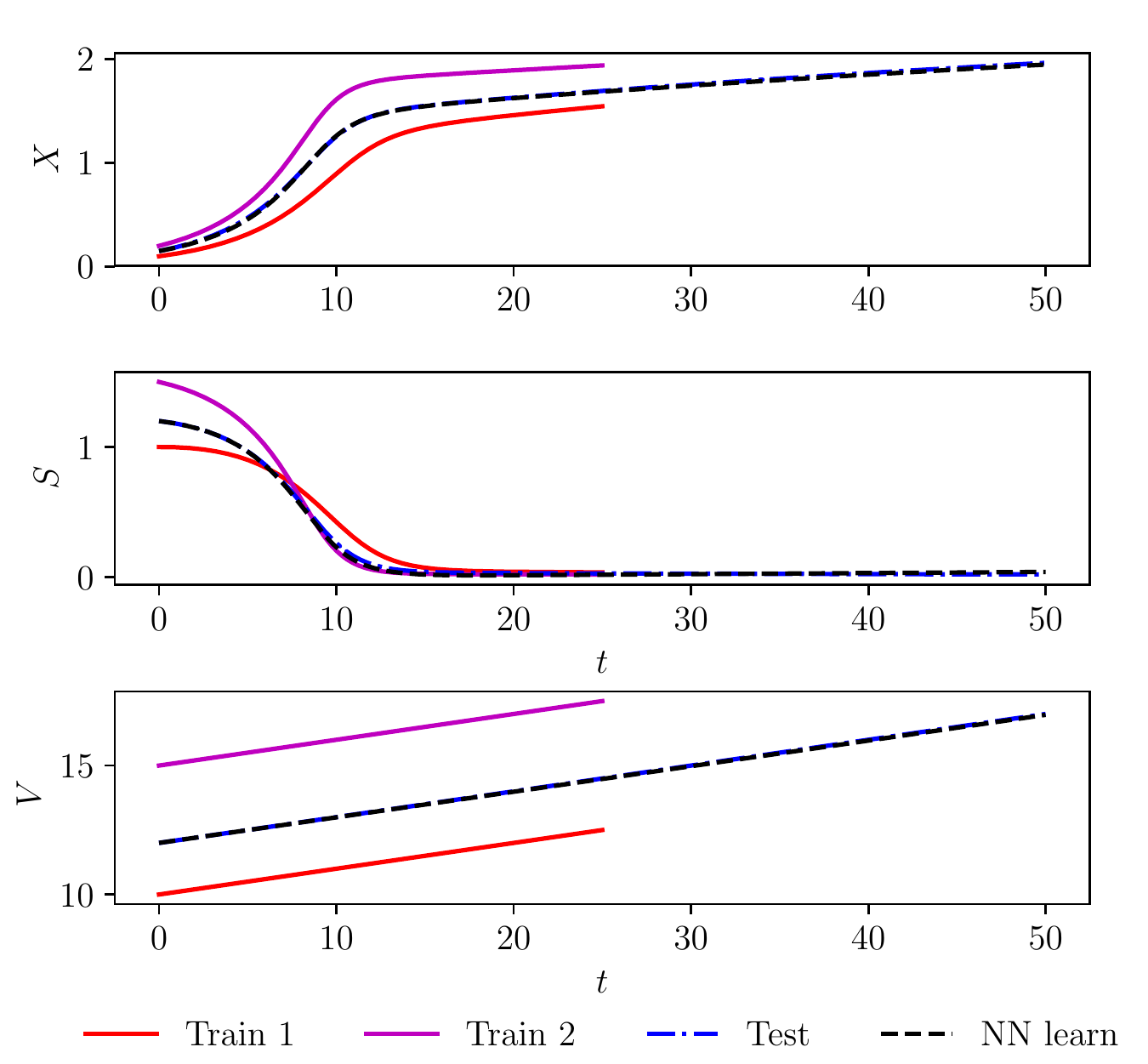}
        \caption{} \label{fig:multistepnn_rhs_state_wo_norm}
    \end{subfigure}        
    \begin{subfigure}[t]{0.48\textwidth}
        \centering
        \includegraphics[scale=.42]{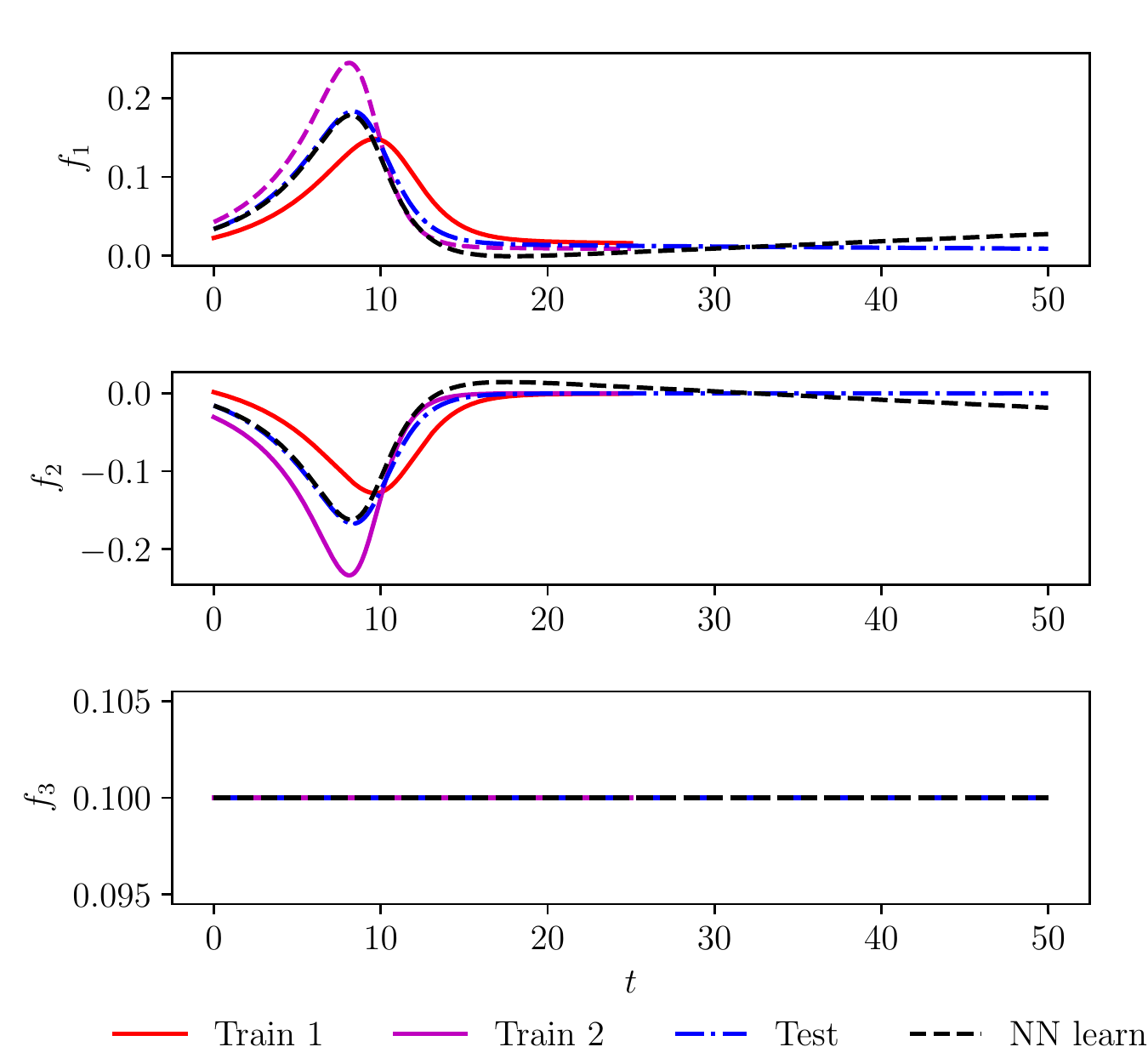}
        \caption{} \label{fig:multistepnn_rhs_rhs_wo_norm}
    \end{subfigure}    
      \caption{Unknown dynamics - Multistep NN without normalization (2 shorter trajectories used for training): a). Solution of FBR model computed from exact dynamics and with learned dynamics using multistep neural networks, b). Right hand side (RHS) of the ODE describing the evolution dynamics.}   \label{multistepnn_rhs_wo_norm}
\end{figure}

Figs. \ref{multistepnn_rhs} and \ref{multistepnn_rhs_wo_norm} 
show the numerical results obtained using the multistep NN model when the  dynamics $f$ are completely unknown. Fig. \ref{fig:multistepnn_rhs_state} shows the training data (in red), test data (in dashed blue) and the prediction of the multistep NN model with learned dynamics $f^{NN}$ (in dashed black) of the states $X(t), S(t)$ and $V(t).$ Note that the initial conditions corresponding to the training data ($X_0 = 0.1,$ $S_0 = 1,$ and $V_0 = 10$) and the test data ($X_0 = 0.12,$ $S_0 = 12,$ and $V_0 = 10$) are different. 

Although the test data (in blue) and the NN-based predictions (in black) match very well for the states $X(t)$ and $V(t)$, they deviate slightly for the state $S(t).$ This is due to the fact that there is an order of difference in magnitude for $X(t)$ and $ S(t)$ compared to that of $V(t).$ This difference in magnitude makes it harder to train a network that learns unknown dynamics. As we show in later plots (Figs. \ref{multistepnn_rhs_wo_norm}) 
this deviation can be remedied by using additional time series from different initial conditions (even if the trajectories are shorter)

Fig. \ref{fig:multistepnn_rhs_rhs} shows the training, learned and test evolutions of the right hand side function $f$ describing the dynamics. We observe that there is a better agreement for the solution than the dynamics (RHS of the equations). It is due to the fact that the solution as a function of time is more smoother than that of the dynamics ($f$).

In Fig. \ref{multistepnn_rhs_wo_norm} we use two sets of training data (red and magenta) obtained with different initial conditions ([0.1,1.0, 10.0] and [0.2,1.5,15.0]) but for a shorter time duration (25 sec) than that of the test data obtained with initial conditions ([0.15, 1.2, 12.0]) and NN-based predictions for a longer time duration(50 sec). We do that to highlight the neural network's ability to learn the correct physics so that it can be used to simulate the dynamics beyond the time interval used for training. Furthermore, we see that the addition of a second set of data slightly improves the accuracy of the predicted state vector (Fig. \ref{fig:multistepnn_rhs_state_wo_norm}) as well as the accuracy of the predicted dynamics (Fig. \ref{fig:multistepnn_rhs_rhs_wo_norm}). 
%

\subsection{Discrete time - constitutive relation}

\begin{figure}[h!]
    \centering
    \begin{subfigure}[t]{0.32\textwidth}
        \centering
        \includegraphics[scale=.32]{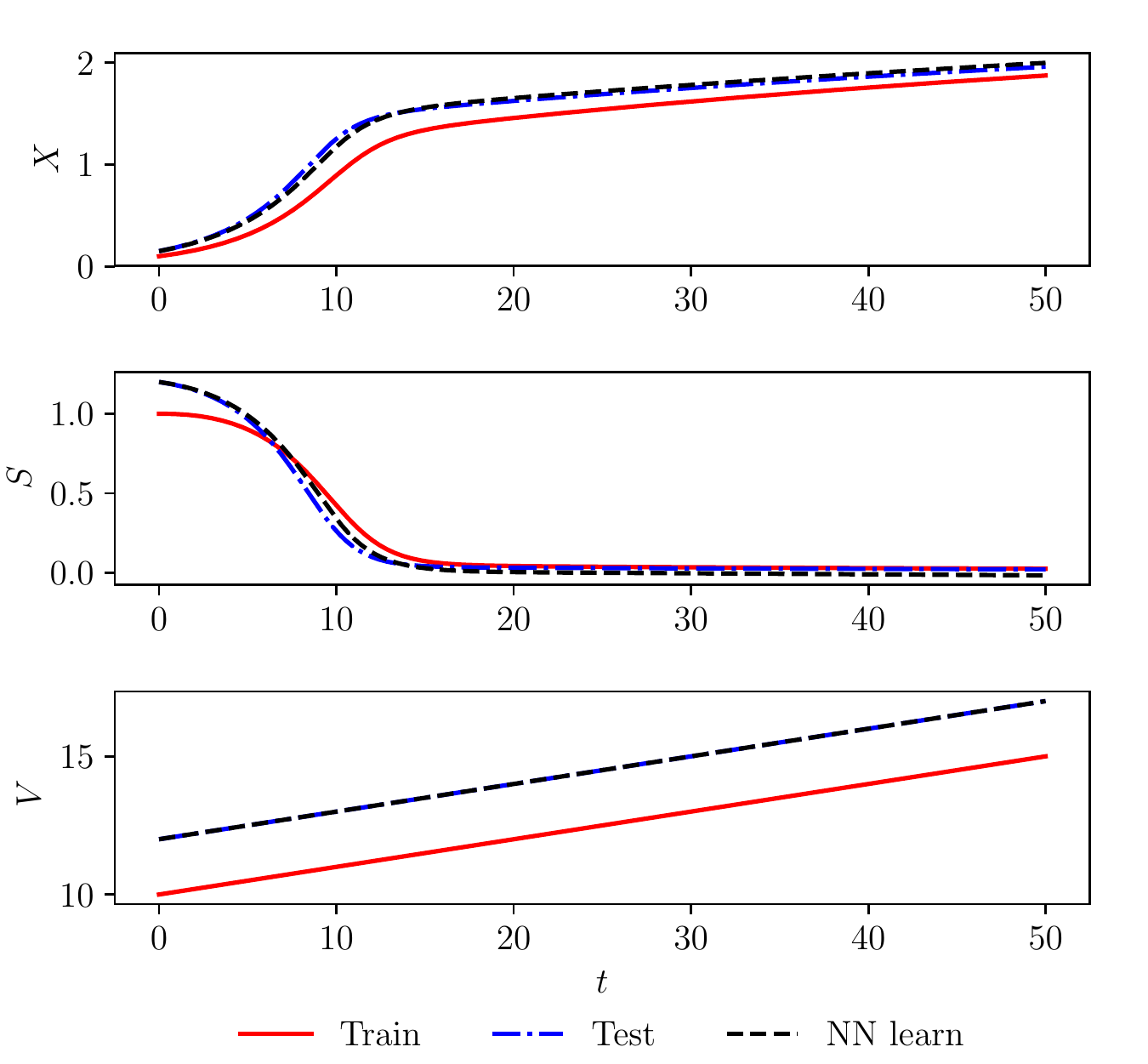}
        \caption{} \label{fig:multistepnn_mu_state}
    \end{subfigure}        
    \begin{subfigure}[t]{0.32\textwidth}
        \centering
        \includegraphics[scale=.35]{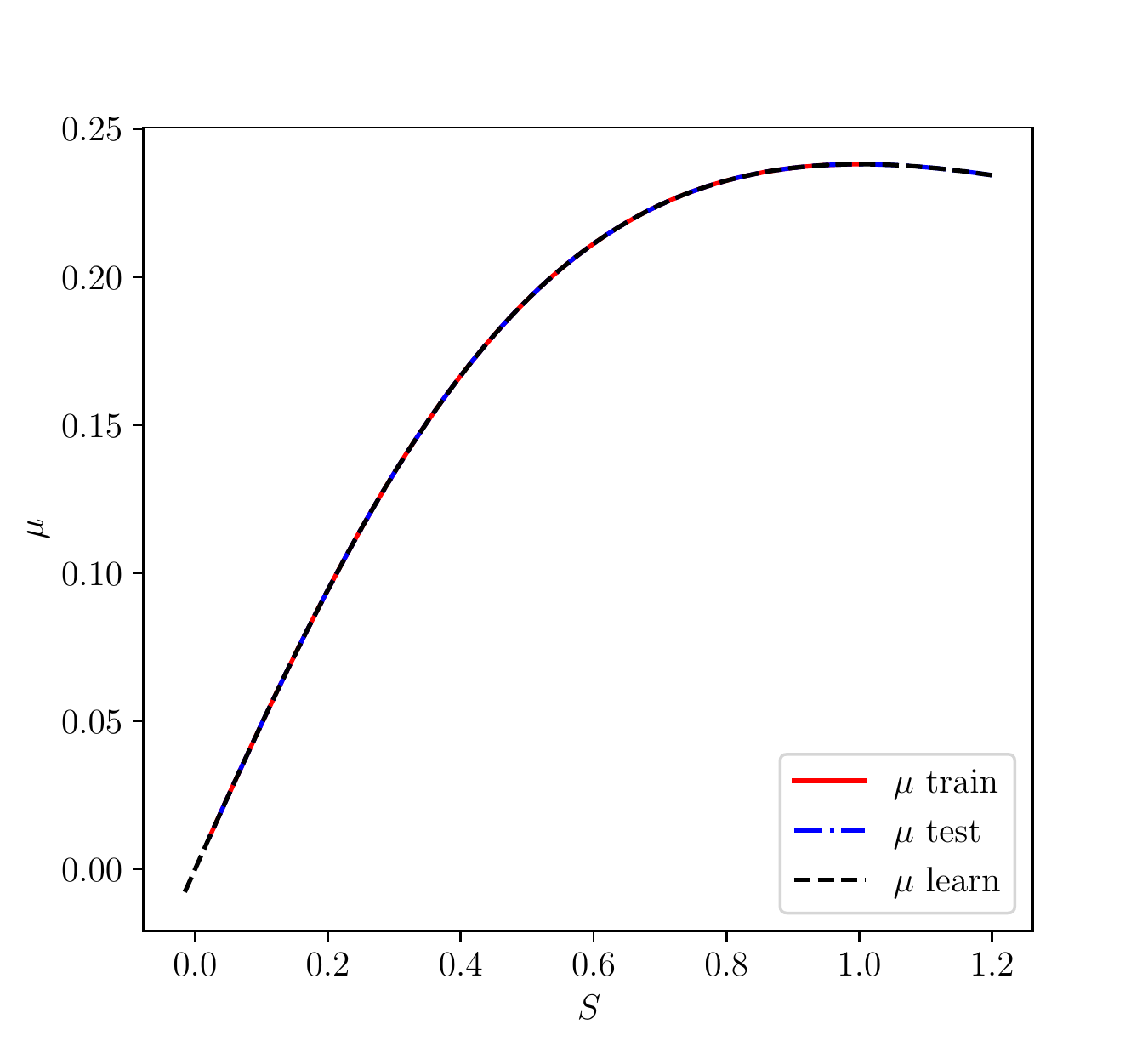}
        \caption{} \label{fig:multistepnn_mu_vs_S}
    \end{subfigure}    
    \begin{subfigure}[t]{0.32\textwidth}
        \centering
        \includegraphics[scale=.35]{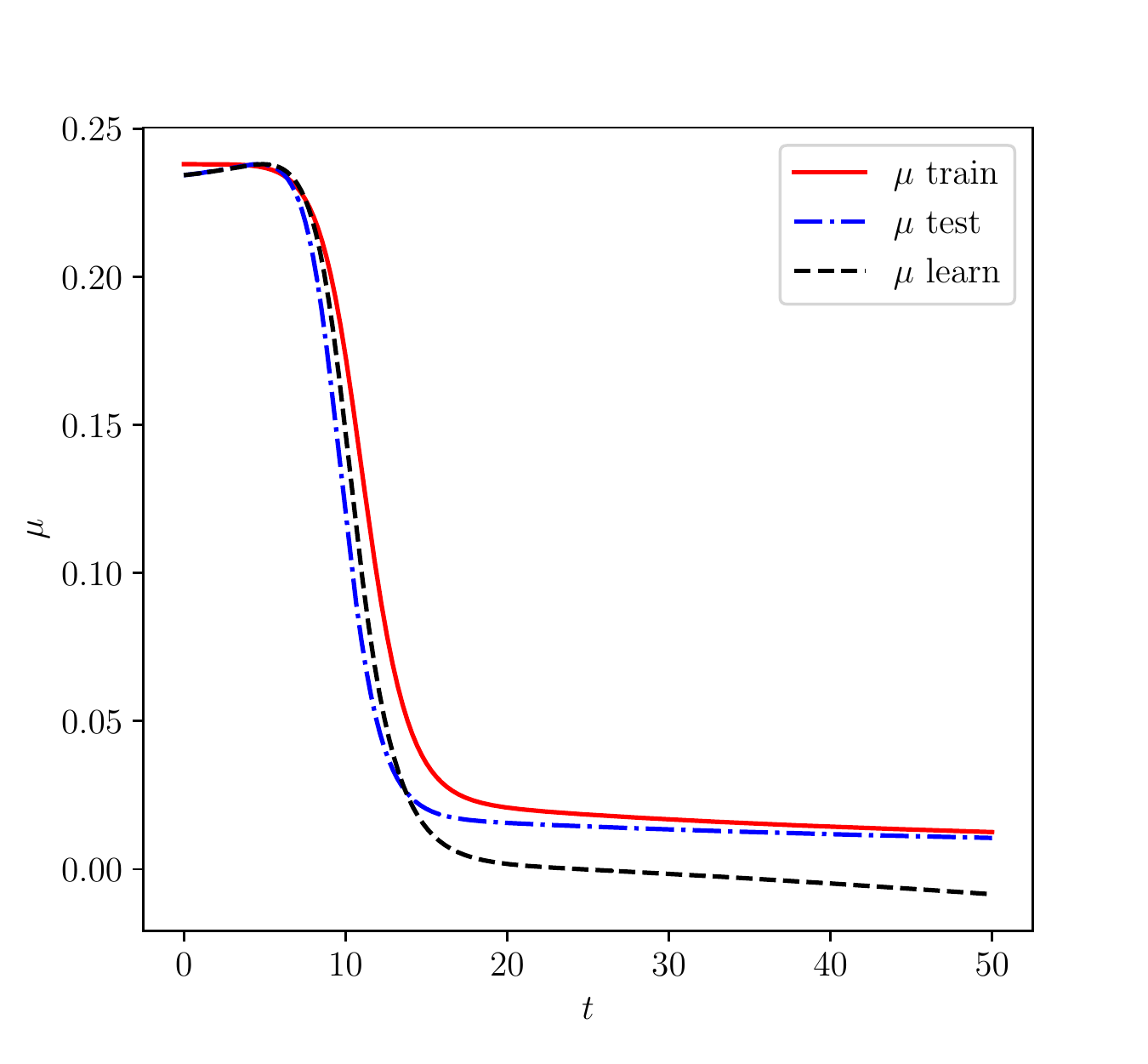}
        \caption{} \label{fig:multistepnn_mu_vs_t}
    \end{subfigure}    
      \caption{constitutive relation - Multistep NN (1 trajectory used for training): a). Solution of FBR model computed from exact dynamics and with learned dynamics using multistep neural networks, b). Coefficient $\mu(t)$ vs state $S(t)$, c). Coefficient $\mu(t)$ vs time $t$.}   \label{multistepnn_mu}
\end{figure}

In Fig. \ref{multistepnn_mu} we show the numerical results obtained with the multistep NN model when the equations governing the dynamics $f$ are known but the constitutive relation  $\mu(y)$ is unknown. Here we model $\mu(y)$ as a neural network that takes the state vector as input and outputs $\mu^{NN}(y)$. 

Fig. \ref{fig:multistepnn_mu_state} shows the training data (in red), test data (in dashed blue) and predictions of the multistep NN model (in dashed black) of the states $X(t), S(t), V(t)$. Note that the initial conditions of the test and NN-learned dynamics are different from those of the training data. Since we have more information about the system, namely the equations governing dynamics, the multistep NN model works very well and its predictions (in dashed black) match well with the test data (dashed blue). Figs. \ref{fig:multistepnn_mu_vs_S} and \ref{fig:multistepnn_mu_vs_t} show the learned constitutive relation  $\mu$ as a function of the state $S$ and time $t$ respectively. These are  compared with the ground truth constitutive relation values  from the Haldane model. There is good agreement between the learned constitutive relation  $\mu^{NN}(S)$ with the Haldane model \eqref{eq:haldane}. 

\begin{figure}[h!]
    \centering
    \begin{subfigure}[t]{0.32\textwidth}
        \centering
        \includegraphics[scale=.32]{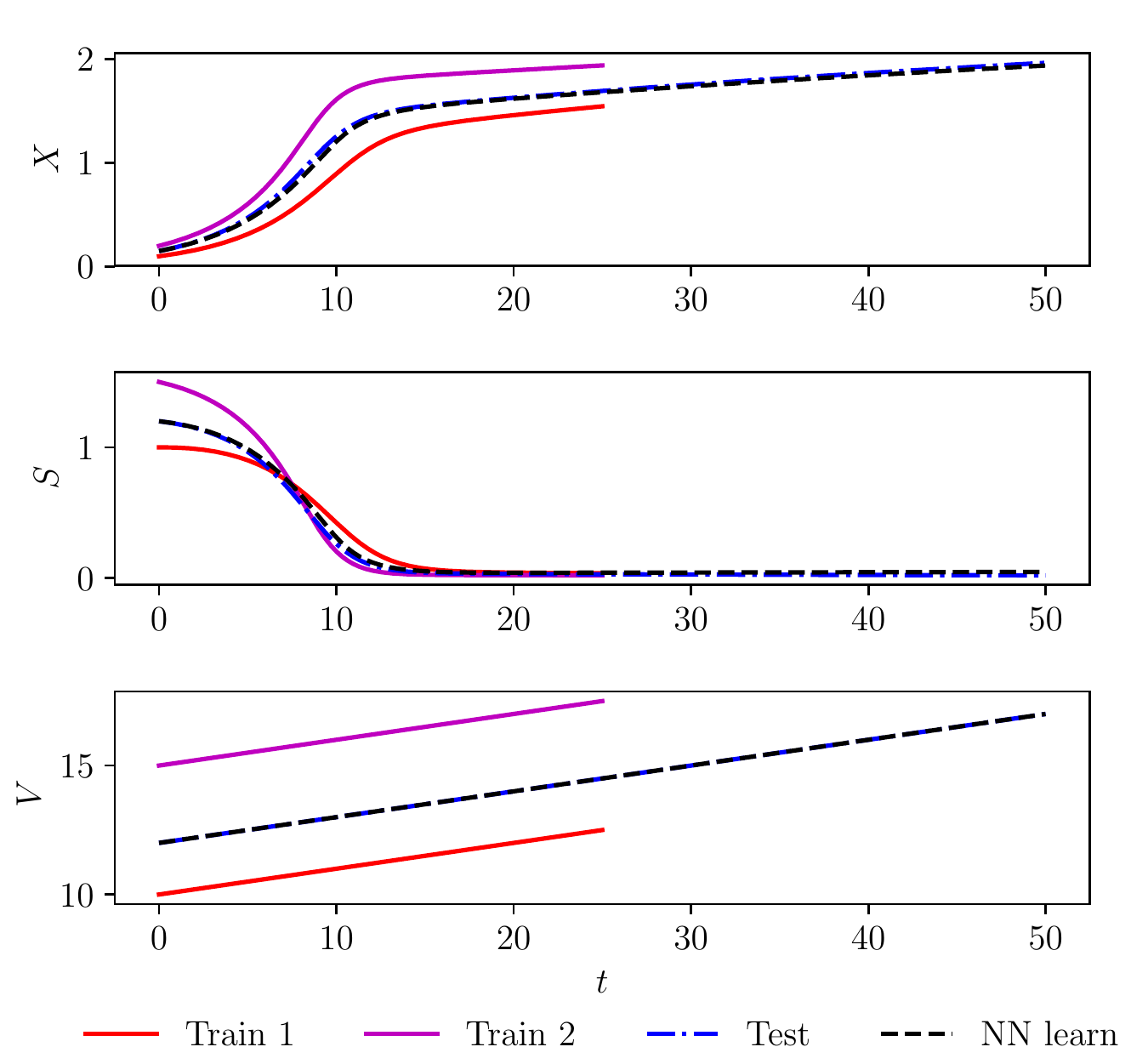}
        \caption{} \label{fig:multistepnn_mu_state_2}
    \end{subfigure}        
    \begin{subfigure}[t]{0.32\textwidth}
        \centering
        \includegraphics[scale=.35]{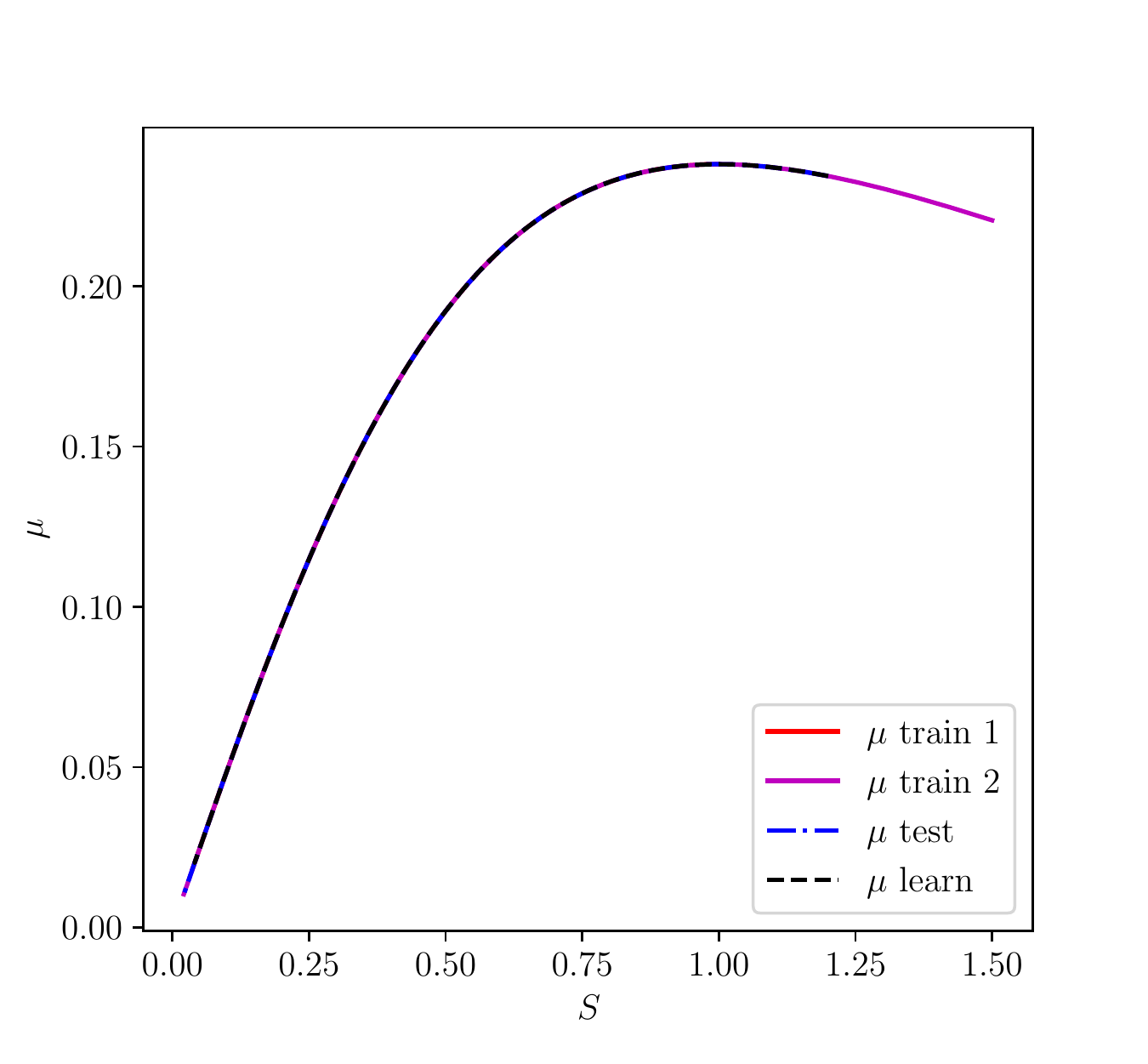}
        \caption{} \label{fig:multistepnn_mu_vs_S_2}
    \end{subfigure}    
    \begin{subfigure}[t]{0.32\textwidth}
        \centering
        \includegraphics[scale=.35]{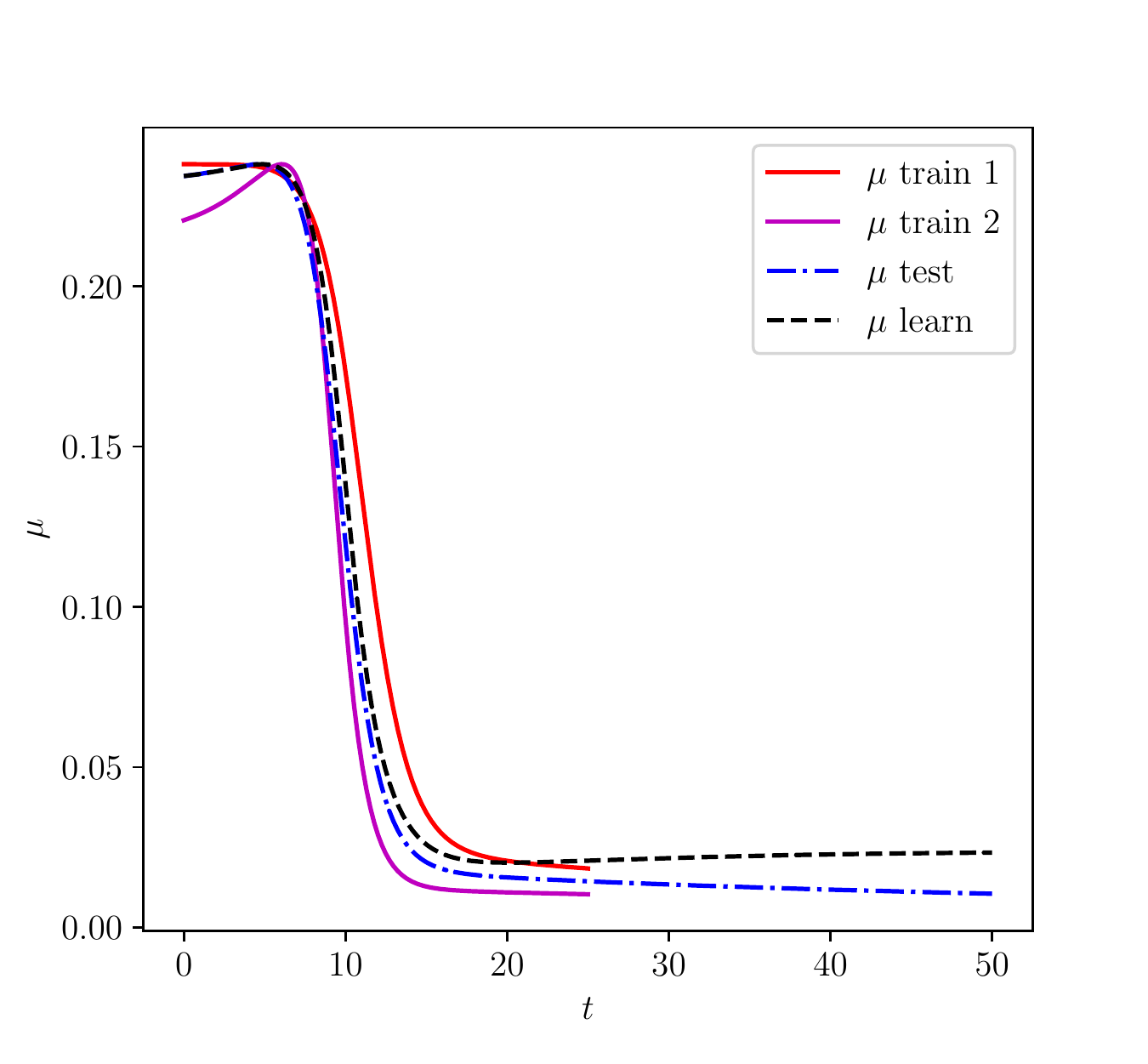}
        \caption{} \label{fig:multistepnn_mu_vs_t_2}
    \end{subfigure}    
      \caption{constitutive relation - Multistep NN (2 shorter trajectories used for training): a). Solution of FBR model computed from exact dynamics and with learned dynamics using multistep neural networks, b). Coefficient $\mu(t)$ vs state $S(t)$, c). Coefficient $\mu(t)$ vs time $t$.}   \label{multistepnn_mu_2}
\end{figure}

 Fig. \ref{multistepnn_mu_2} shows similar plots as those in Fig. \ref{multistepnn_mu} with two sets of training data (shown in red and magenta) which extend to shorter time (25 sec) compared to the length of the test data and learned predictions (50 sec). Once again we can observe very good agreement between the NN-based predictions and the test data (Fig. \ref{fig:multistepnn_mu_state_2}). The same applies to the agreement between the learned constitutive relation and ground truth from the Haldane model (Figs. \ref{fig:multistepnn_mu_vs_S_2} and \ref{fig:multistepnn_mu_vs_t_2}).   

\subsection{Continuous time - Unknown dynamics}

\begin{figure}[h!]
    \centering
    \begin{subfigure}[t]{0.48\textwidth}
        \centering
        \includegraphics[scale=.42]{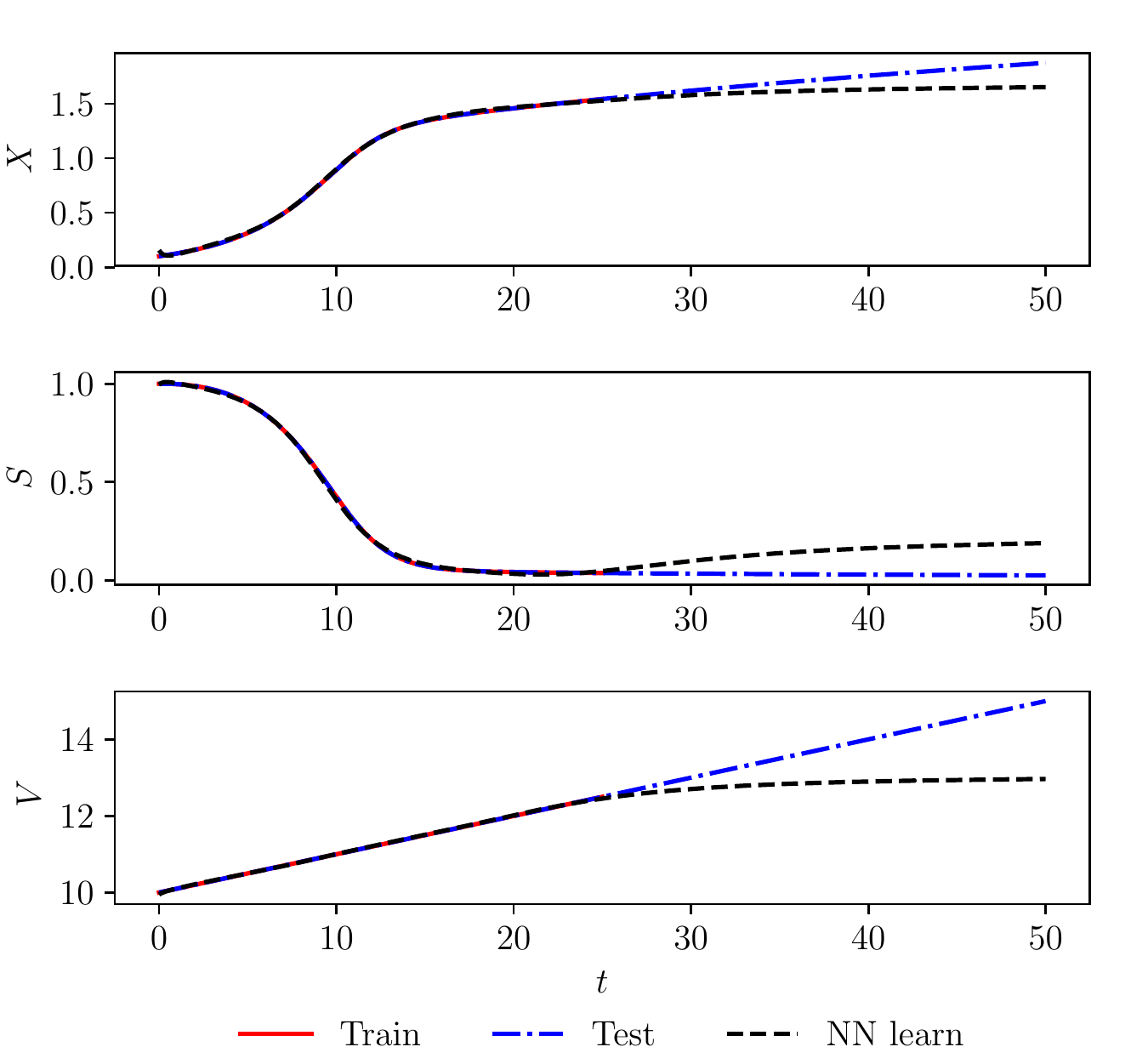}
        \caption{} \label{fig:contpinns_f_state}
    \end{subfigure}        
    \begin{subfigure}[t]{0.48\textwidth}
        \centering
        \includegraphics[scale=.42]{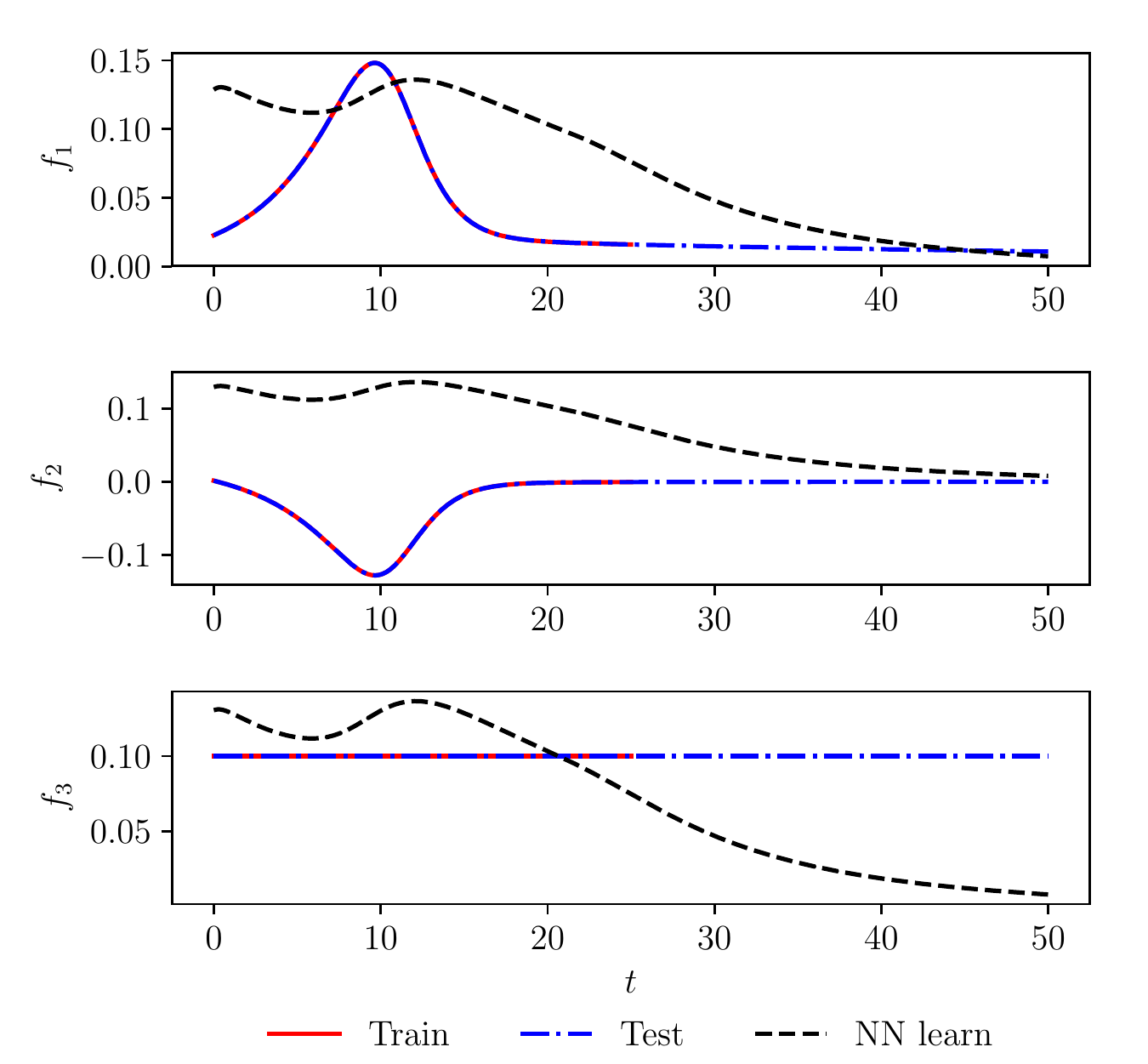}
        \caption{} \label{fig:contpinns_f_rhs}   
    \end{subfigure}    
      \caption{Unknown dynamics - Continuous time (1 shorter trajectory with same initial condition is used for training): a). Solution of FBR model computed from exact dynamics and with learned dynamics using continuous time neural networks, b). Right hand side (RHS) of the ODE describing the evolution dynamics.}   \label{contpinns_f}
\end{figure}

\begin{figure}[h!]
    \centering
    \begin{subfigure}[t]{0.48\textwidth}
        \centering
        \includegraphics[scale=.42]{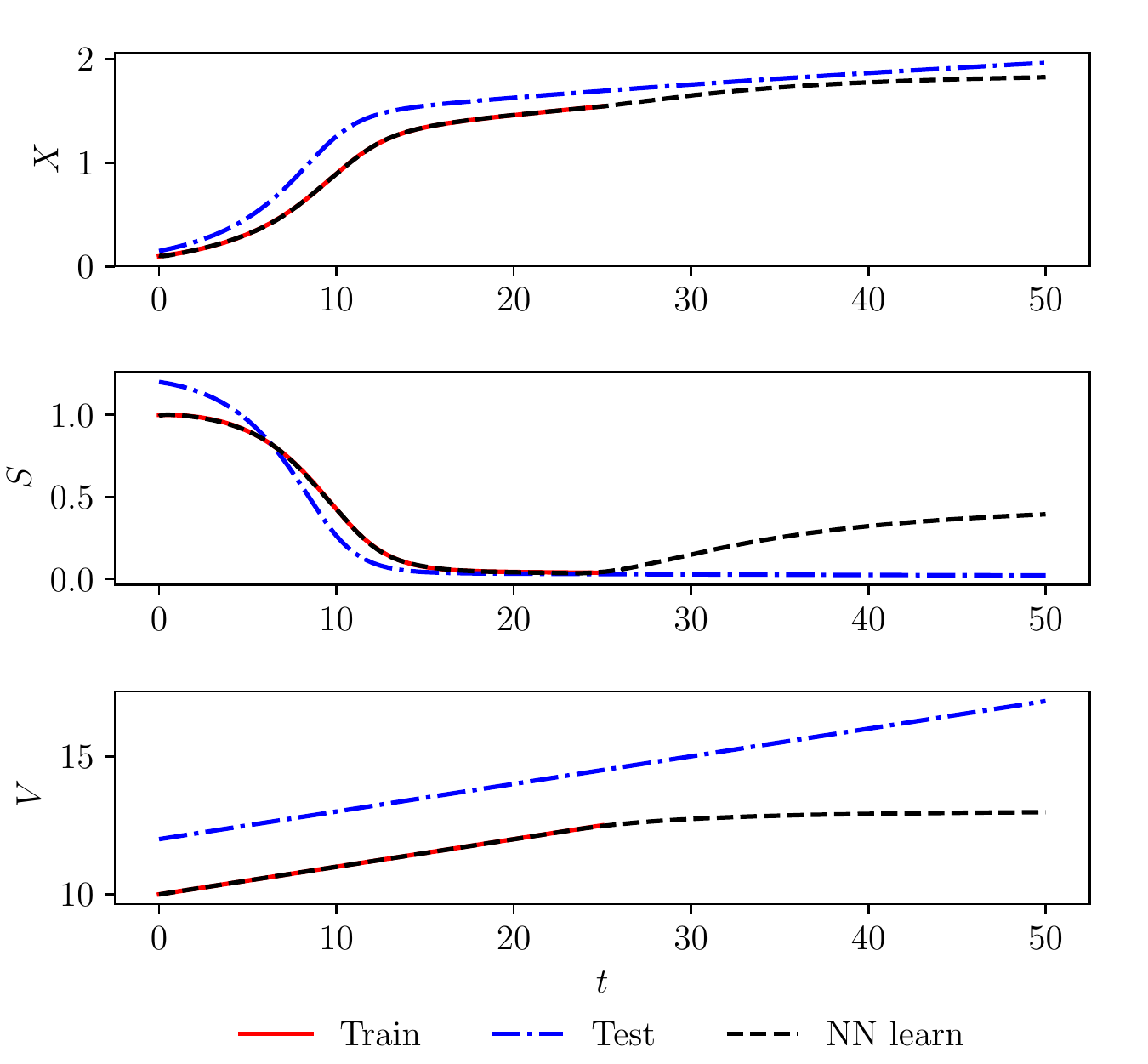}
        \caption{} \label{fig:contpinns_f_state_diff_ics}
    \end{subfigure}        
    \begin{subfigure}[t]{0.48\textwidth}
        \centering
        \includegraphics[scale=.42]{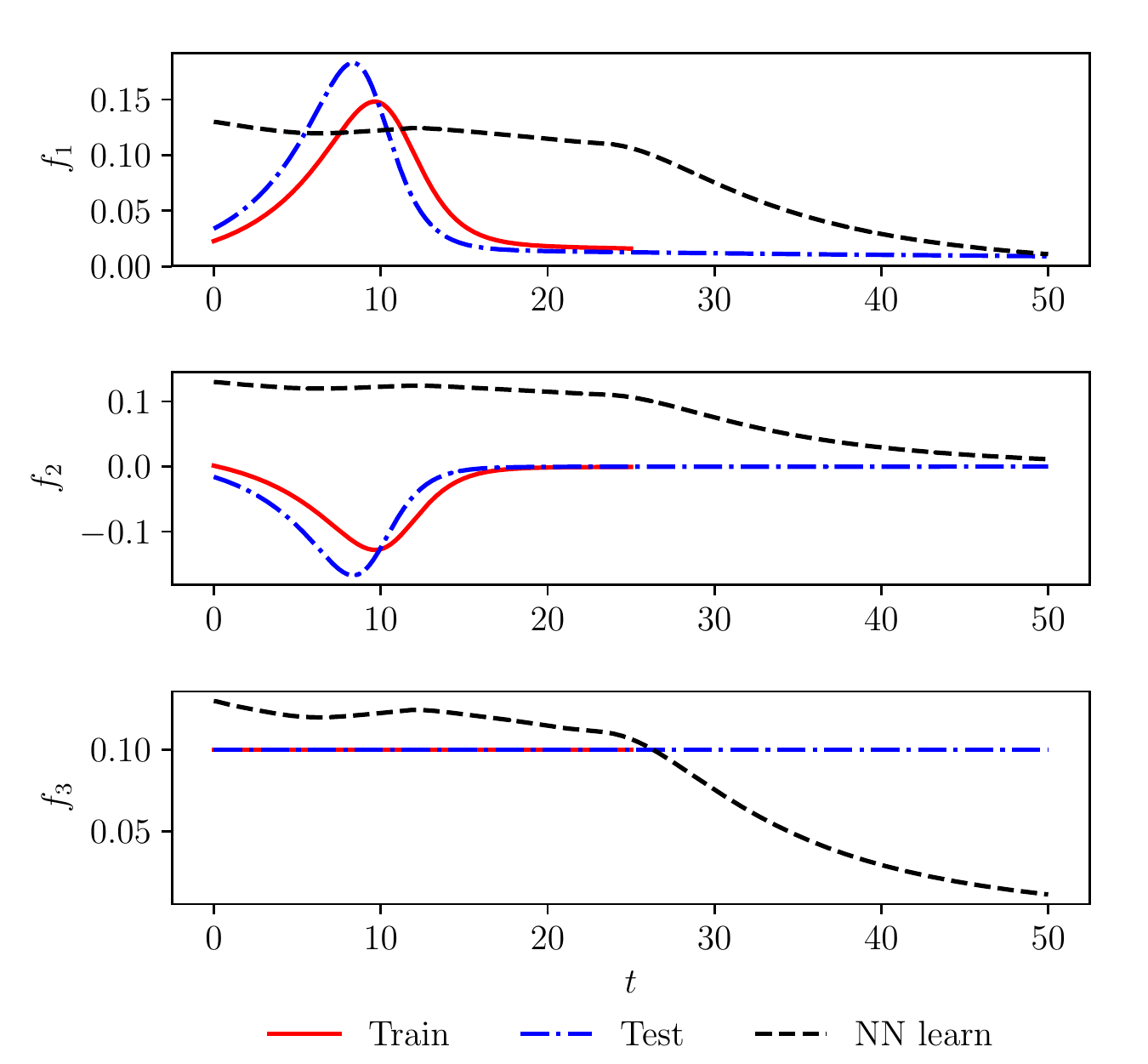}
        \caption{} \label{fig:contpinns_f_rhs_diff_ics}   
    \end{subfigure}    
      \caption{Unknown dynamics - Continuous time (1 shorter trajectory with different initial condition is used for training): a). Solution of FBR model computed from exact dynamics and with learned dynamics using continuous time neural networks, b). Right hand side (RHS) of the ODE describing the evolution dynamics.}   \label{contpinns_f_diff_ics}
\end{figure}

Fig. \ref{contpinns_f} shows the results from continuous physics-informed neural network method for unknown dynamics case described in section \ref{sec:contpinn_rhs}. The figure shows the system state training data in red and learned solution from $y^{NN}$ in dashed blue. Here the initial conditions ($[0.1, 1.0, 10.0]$) for the system dynamics is same for both training dynamics and NN learned dynamics, however the time duration of the training dynamics (25 sec) is shorter than that of the test and NN learned dynamics. Here, as observed in Fig. \ref{contpinns_f} the NN $y^{NN}$ acts as an interpolation function, that is the NN learned solution and the RHS describing the dynamics ($f$) agree very well with the test solution with in the interpolation range (from 0 to 25 sec) and deviates in the extrapolation range (from 25 to 50 sec). This approach also doesn't work for test data with different initial conditions than those of the training data. Fig. \ref{contpinns_f_diff_ics} shows the results obtained with this approach where the initial conditions of the training data ($[0.1, 1.0, 10.0]$) are different from those of the test data ($[0.15, 1.2, 12.0]$).  Fig. \ref{fig:contpinns_f_state_diff_ics} shows the system states with training data in red, test solution in dashed blue and learned solution in dashed black. We can clearly see that the NN learned solution from $y^{NN}$ matches with the training data instead of the test data confirming again that this approach is good for interpolation alone. Fig. \ref{fig:contpinns_f_rhs_diff_ics} shows the similar plot for the RHS function $f$.

\begin{figure}[h!]
    \centering
    \begin{subfigure}[t]{0.48\textwidth}
        \centering
        \includegraphics[scale=.42]{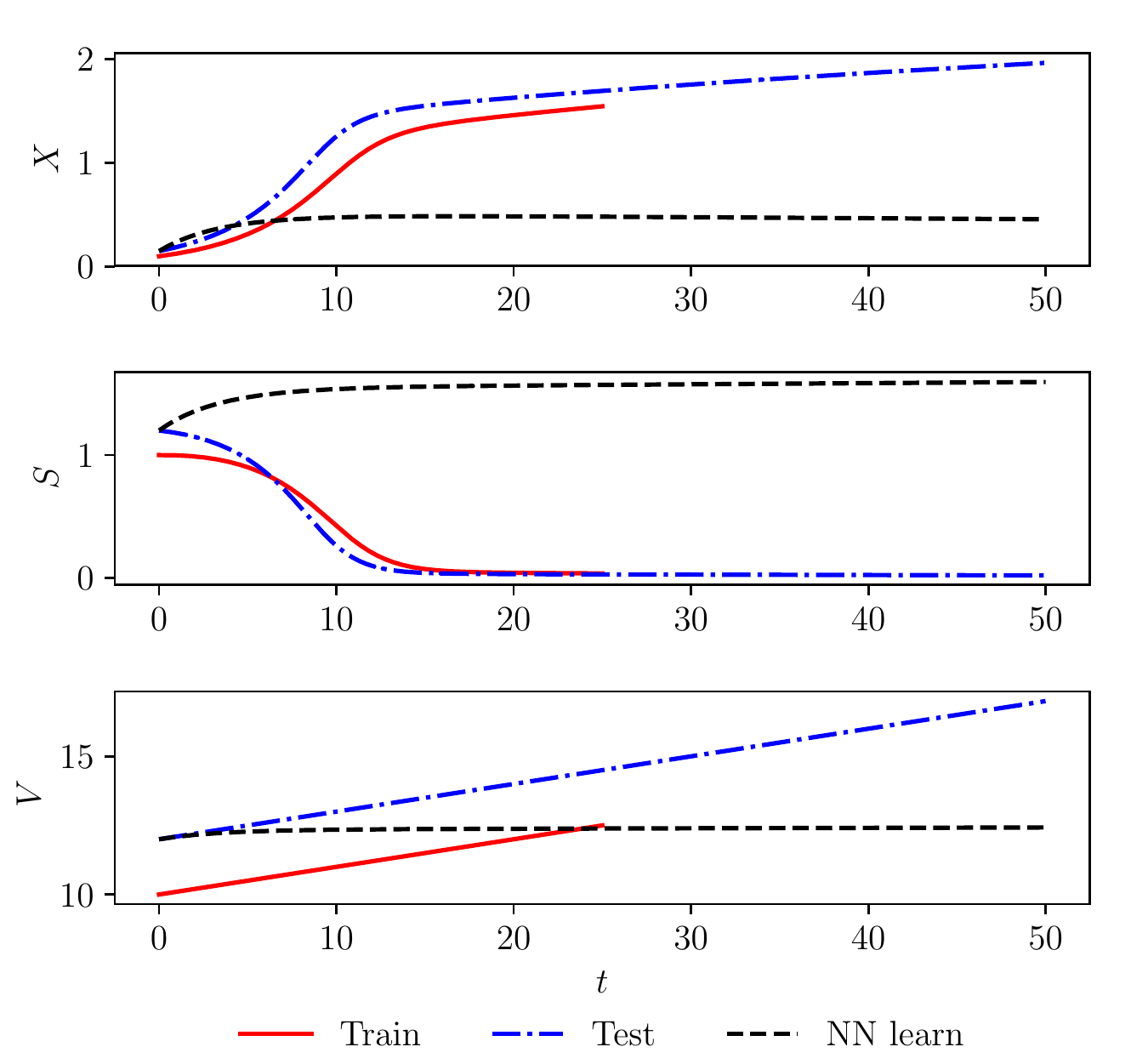}
        \caption{} \label{fig:contpinns_f_state_diff_ics_solve}
    \end{subfigure}        
    \begin{subfigure}[t]{0.48\textwidth}
        \centering
        \includegraphics[scale=.42]{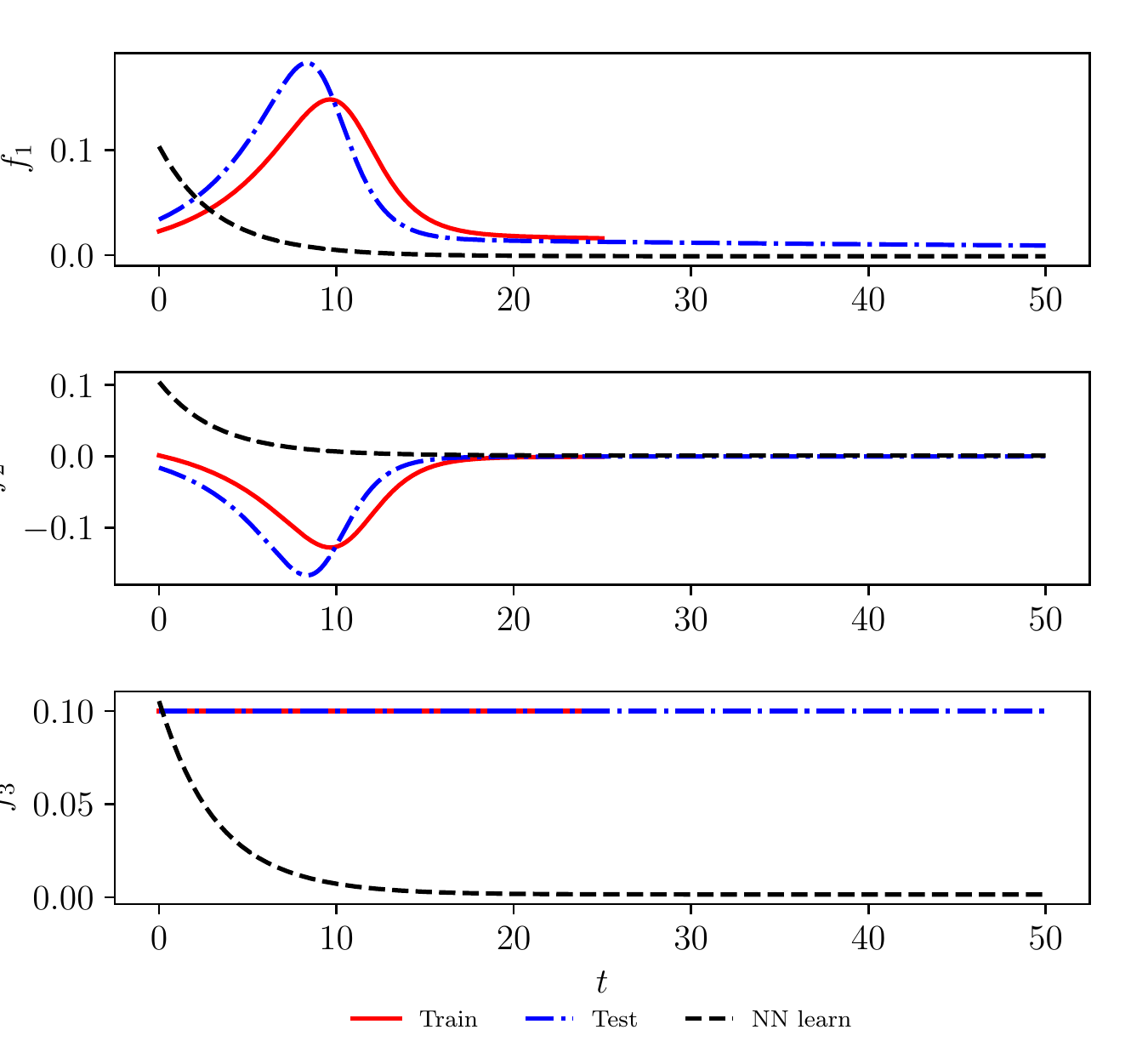}
        \caption{} \label{fig:contpinns_f_rhs_diff_ics_solve}   
    \end{subfigure}    
      \caption{Unknown dynamics - Continuous time (1 shorter trajectory with different initial condition is used for training): a). Solution of FBR model computed from exact dynamics and with learned dynamics using continuous time neural networks, b). Right hand side (RHS) of the ODE describing the evolution dynamics.}   \label{contpinns_f_diff_ics_solve}
\end{figure}

\begin{figure}[h!]
    \centering
    \includegraphics[scale=.75]{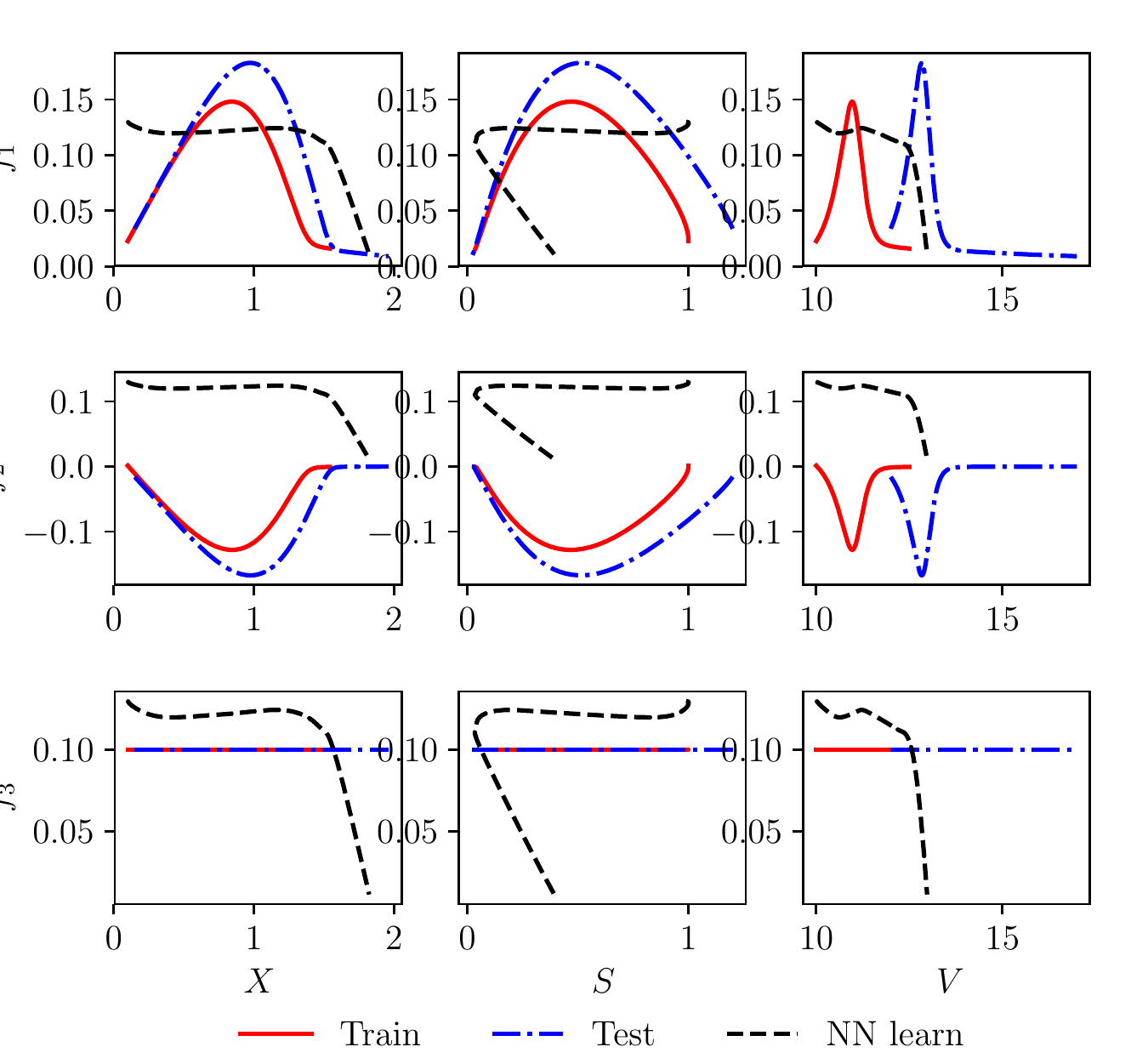}
    \caption{Unknown dynamics - Continuous time: Right hand side $f$ of FBR model computed from exact dynamics and with learned dynamics as a function of the system state}   \label{contpinns_f_vs_state}
\end{figure}

To address this issue we also attempted to train the NN to learn the right hand side function $f(y)$ as a function of the system states $y$ so that one can solve the governing equations with RHS as the learned NN $f^{NN}(y).$ However, the NN $f^{NN}(y)$ obtained with this approach is not accurate and results in rapid error accumulation. Fig. \ref{contpinns_f_diff_ics_solve} shows the results obtained with this approach. Fig. \ref{fig:contpinns_f_state_diff_ics_solve} shows the system states where the learned dynamics in dashed black shows the solution obtained by solving the governing equation using $f^{NN}(y)$ as the right hand side function. Fig. \ref{fig:contpinns_f_rhs_diff_ics_solve} shows $f^{NN}$ (in dashed black) as a function of time. These results clearly show that the continuous physics informed machine learning (CPINN) method does not produce good approximation of the unknown dynamics. This approach requires further investigation and may require additional constraints and a new loss function.
Fig. \ref{contpinns_f_vs_state} shows the rhs ($f$) as a function of the system states. This plot also confirms that $f^{NN}$ (in dashed black) does not approximate the right hand side function accurately. 


%
%

\subsection{Continuous time - constitutive relation}

\begin{figure}[h!]
    \centering
    \begin{subfigure}[t]{0.32\textwidth}
        \centering
        \includegraphics[scale=.32]{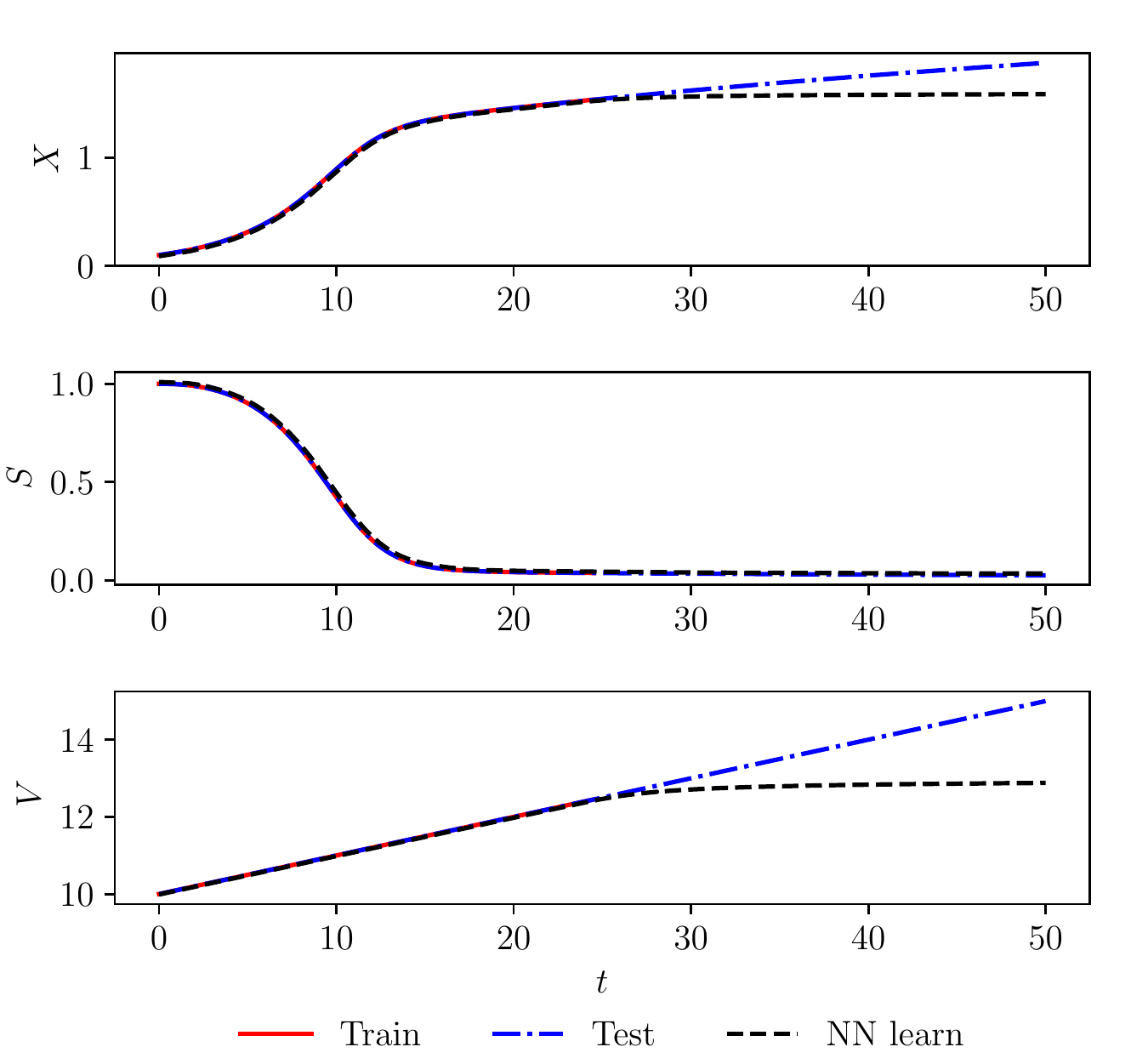}
        \caption{} \label{fig:contpinns_mu_state}
    \end{subfigure}        
    \begin{subfigure}[t]{0.32\textwidth}
        \centering
        \includegraphics[scale=.35]{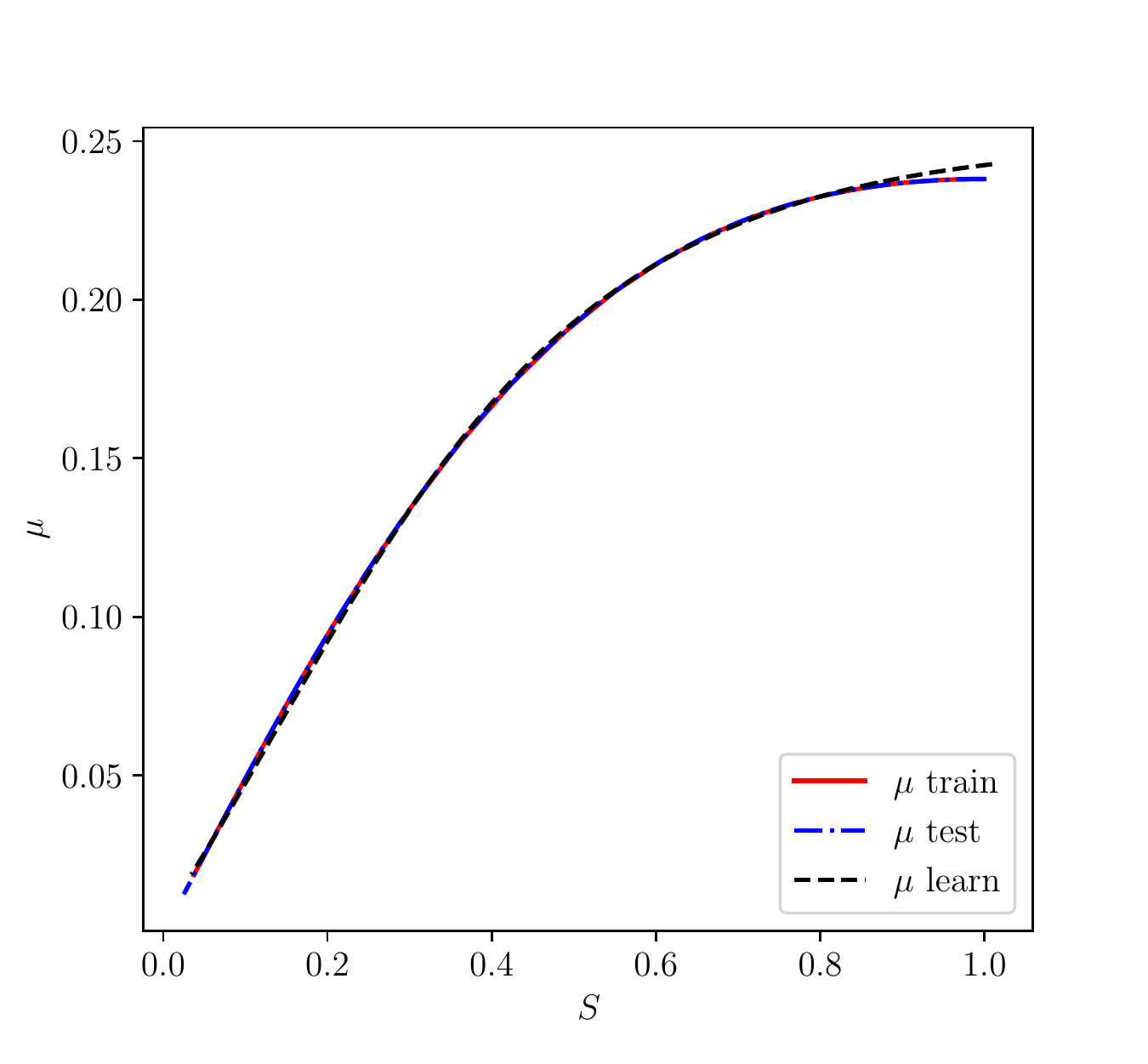}
        \caption{} \label{fig:contpinns_mu_vs_S}
    \end{subfigure}    
    \begin{subfigure}[t]{0.32\textwidth}
        \centering
        \includegraphics[scale=.35]{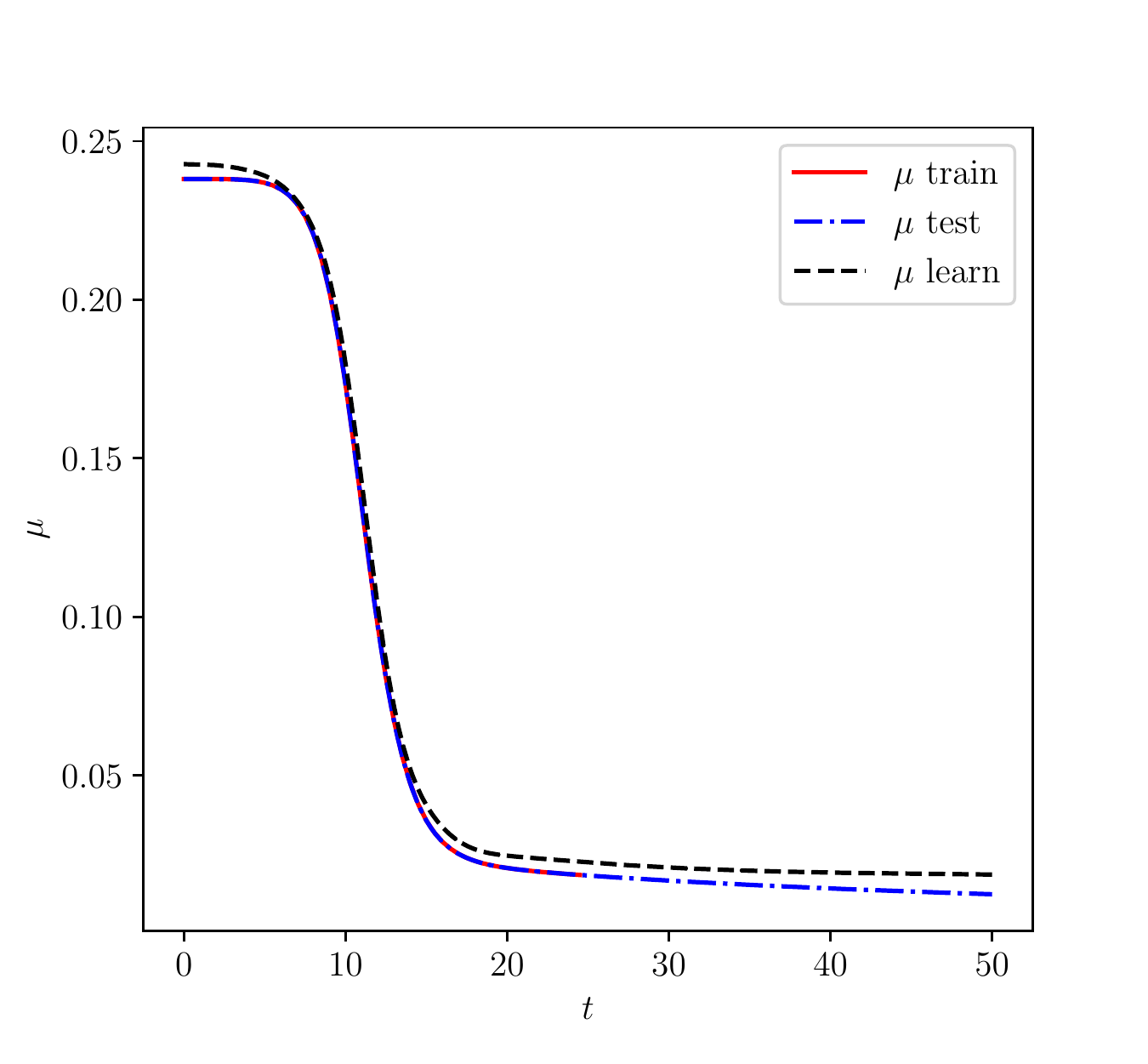}
        \caption{} \label{fig:contpinns_mu_vs_t}
    \end{subfigure}    
      \caption{constitutive relation - Continuous time: a). Solution of FBR model computed from exact dynamics and with learned dynamics, b). Coefficient $\mu(t)$ vs state $S(t)$, c). Coefficient $\mu(t)$ vs time $t$.}   \label{contpinns_mu}
\end{figure}

\begin{figure}[h!]
    \centering
    \begin{subfigure}[t]{0.32\textwidth}
        \centering
        \includegraphics[scale=.33]{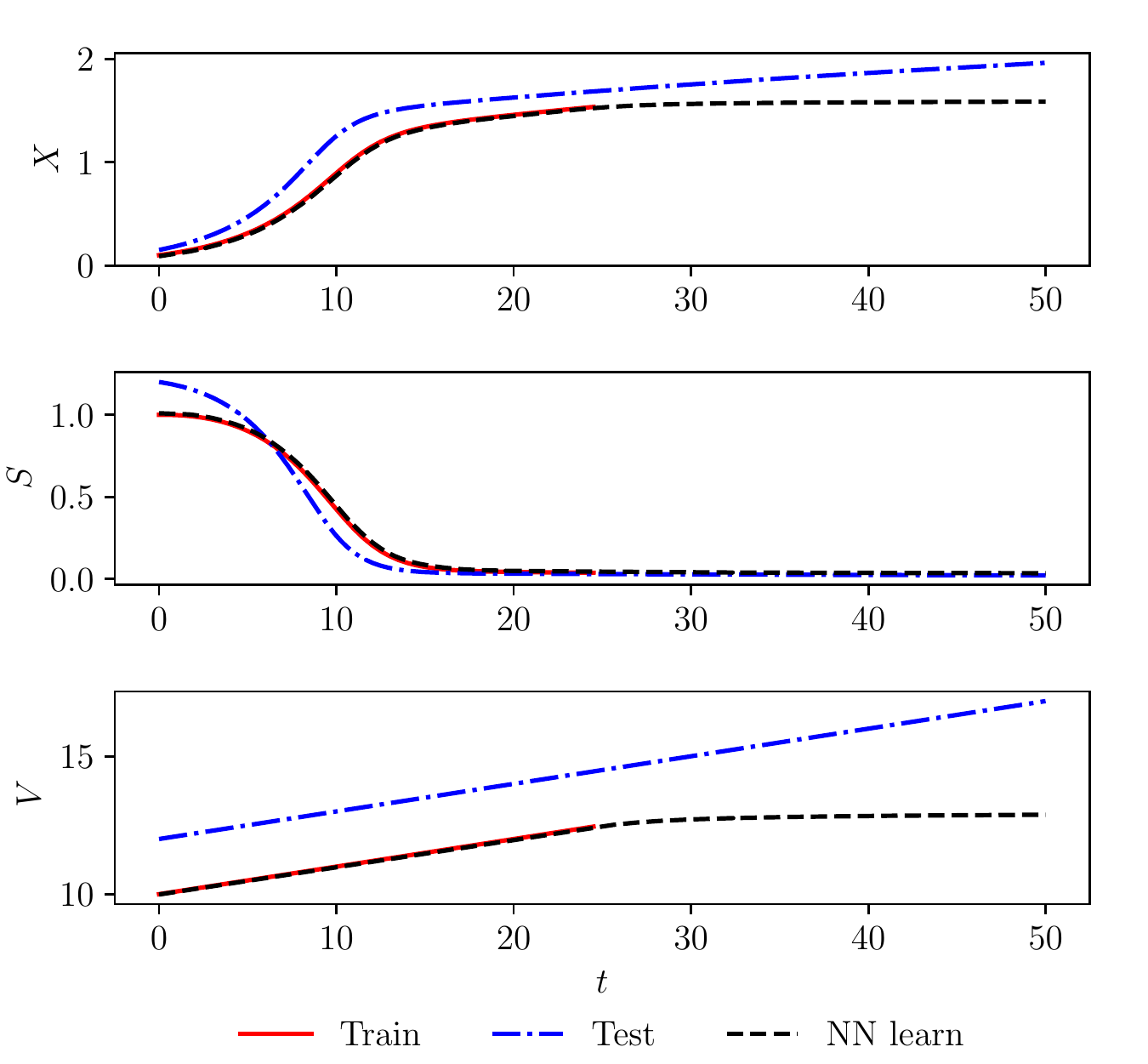}
        \caption{} \label{fig:contpinns_mu_state_diff_ics}
    \end{subfigure}        
    \begin{subfigure}[t]{0.32\textwidth}
        \centering
        \includegraphics[scale=.35]{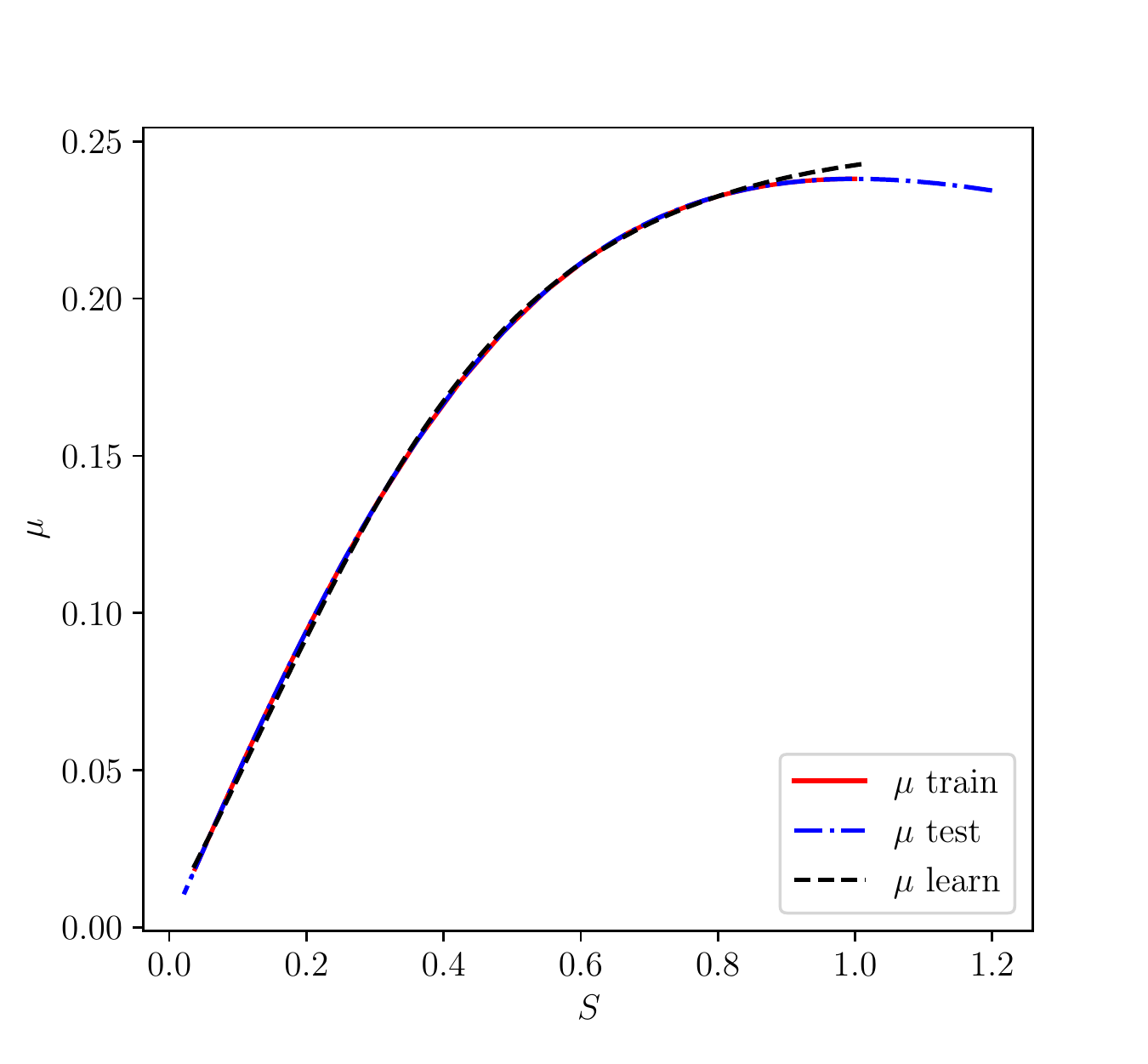}
        \caption{} \label{fig:contpinns_mu_vs_S_diff_ics}
    \end{subfigure}    
    \begin{subfigure}[t]{0.32\textwidth}
        \centering
        \includegraphics[scale=.35]{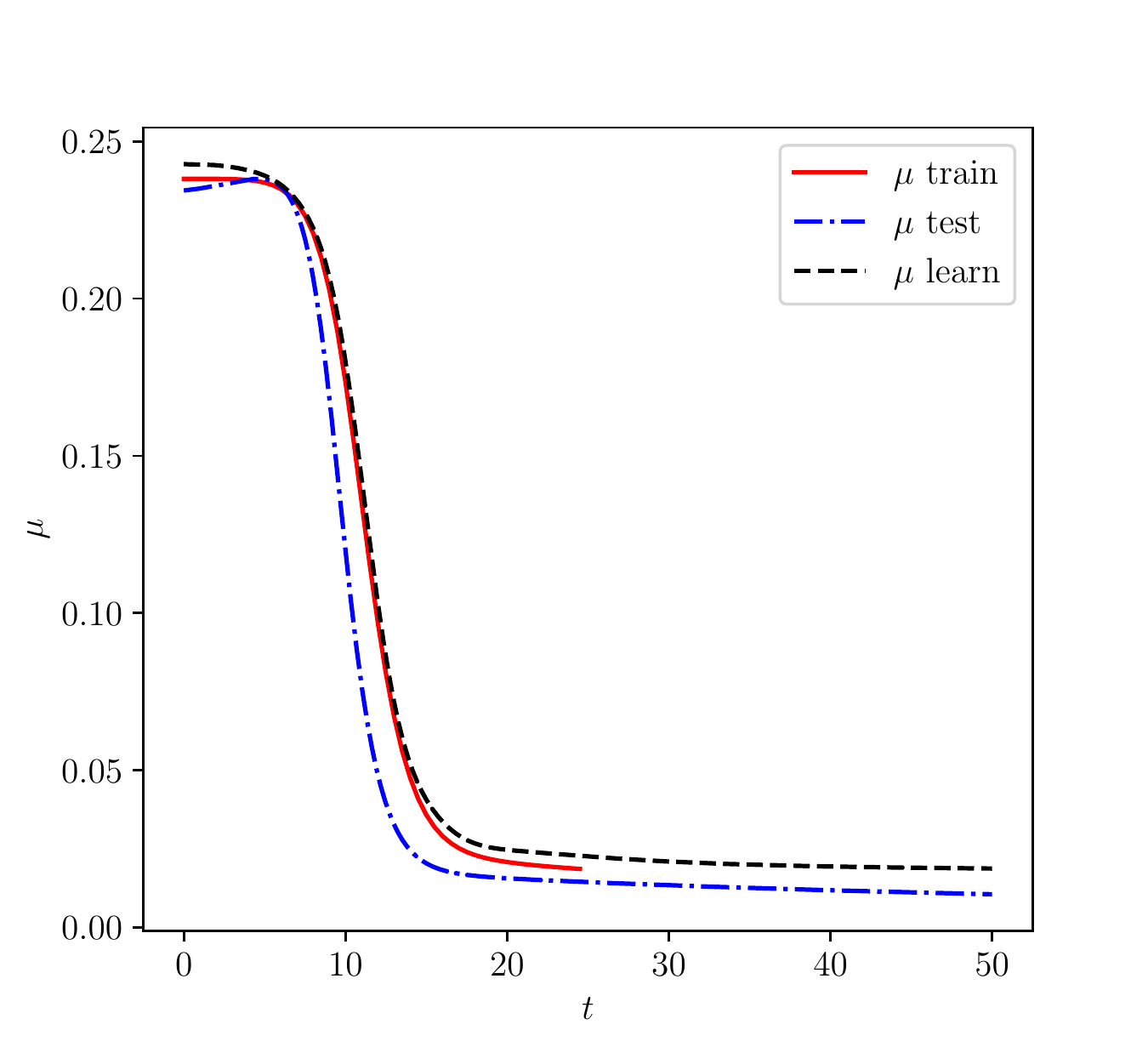}
        \caption{} \label{fig:contpinns_mu_vs_t_diff_ics}
    \end{subfigure}    
      \caption{constitutive relation - Continuous time (1 shorter trajectory and different initial conditions used for training): a). Solution of FBR model computed from exact dynamics and with learned dynamics, b). Coefficient $\mu(t)$ vs state $S(t)$, c). Coefficient $\mu(t)$ vs time $t$.}  \label{contpinns_mu_diff_ics}
\end{figure}

Finally, Fig. \ref{contpinns_mu} shows the results from continuous physics-informed neural network method described in section \ref{sec:contpinn_mu} for learning the constitutive relation  $\mu$.  We can see that the NN learned solution (Fig. \ref{fig:contpinns_mu_state}), and learned constitutive relation  (Figs. \ref{fig:contpinns_mu_vs_S} and \ref{fig:contpinns_mu_vs_t}) agree very well with those of test dynamics and true constitutive relation  respectively. Here, we use the NN $y^{NN}$ which is good for interpolation to obtain the NN learned solution. As we can see from Figs. \ref{fig:contpinns_mu_vs_S} and \ref{fig:contpinns_mu_vs_t}, the learned constitutive relation process using the NN model $\mu^{NN}$ matches very well with that of the true constitutive relation process $\mu$. Fig. \ref{contpinns_mu_diff_ics} shows similar plots with different initial conditions for training and test data. In Fig. \ref{fig:contpinns_mu_vs_S_diff_ics}, the solution from $y^{NN}$ is shown as dashed black line. We can clearly see that this solution follows the training data trajectory (in solid red) instead of test data trajectory (in dashed blue). However, we can see in Fig. \ref{fig:contpinns_mu_vs_S_diff_ics} that the constitutive relation $\mu(S)$ is approximated very well as a function of the state $S(t)$. 

\begin{figure}[h!]
    \centering
    \begin{subfigure}[t]{0.32\textwidth}
        \centering
        \includegraphics[scale=.33]{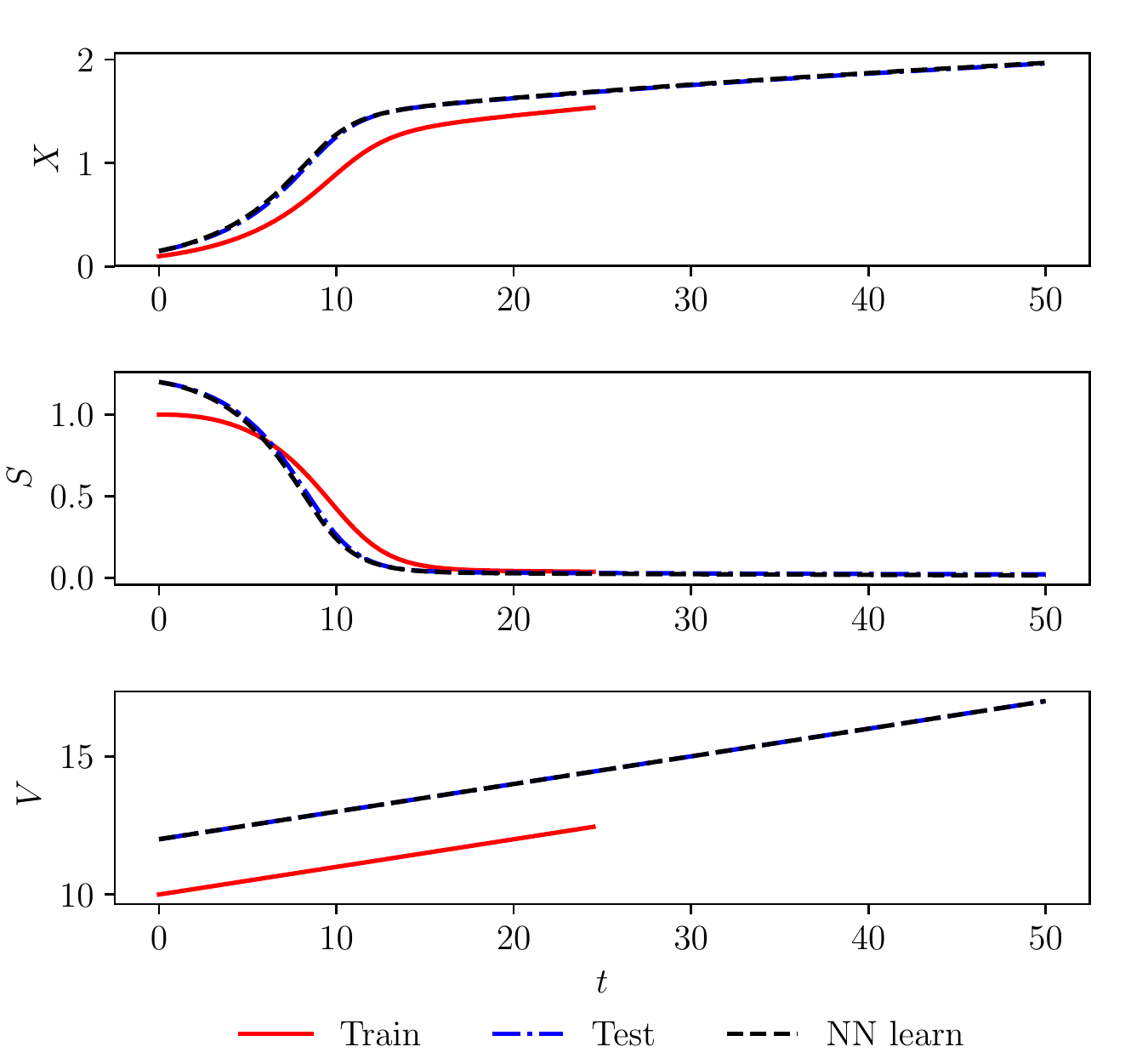}
        \caption{} \label{fig:contpinns_mu_state_diff_ics_solve}
    \end{subfigure}        
    \begin{subfigure}[t]{0.32\textwidth}
        \centering
        \includegraphics[scale=.35]{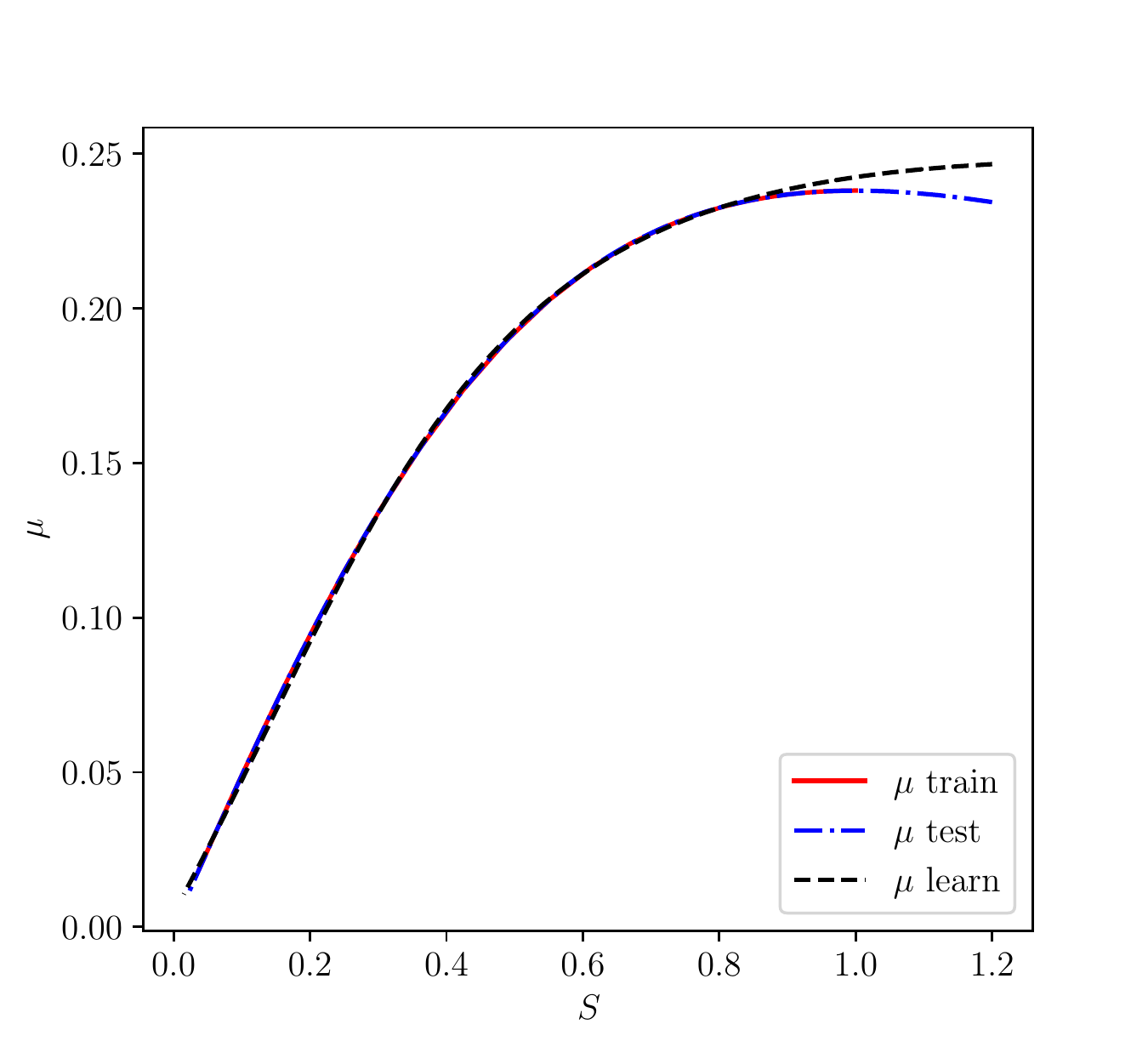}
        \caption{} \label{fig:contpinns_mu_vs_S_diff_ics_solve}
    \end{subfigure}    
    \begin{subfigure}[t]{0.32\textwidth}
        \centering
        \includegraphics[scale=.35]{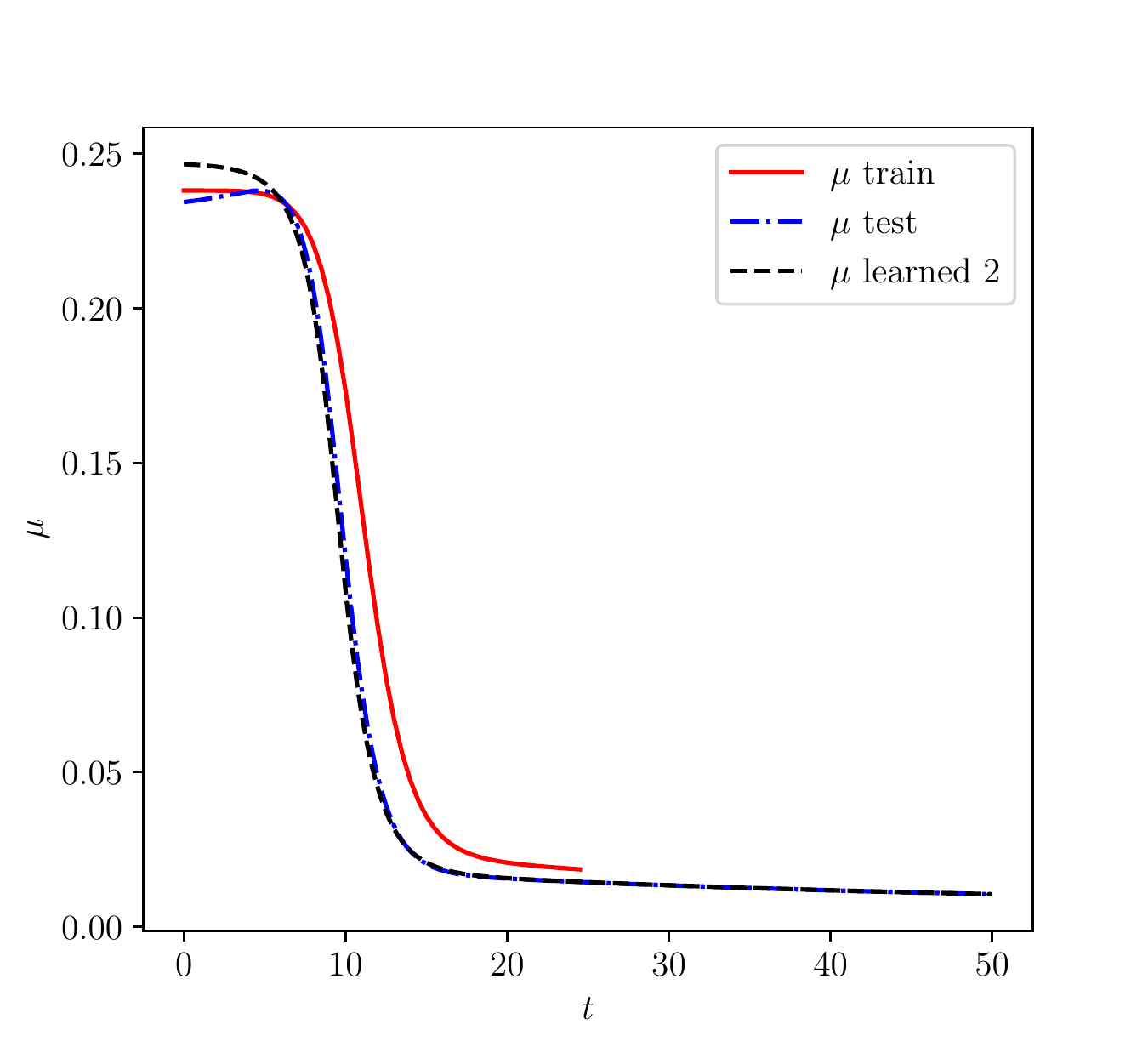}
        \caption{} \label{fig:contpinns_mu_vs_t_diff_ics_solve}
    \end{subfigure}    
      \caption{constitutive relation - Continuous time (1 shorter trajectory and different initial conditions used for training. Solution obtained by solving governing equations with $\mu^{NN}$ as constitutive relation): a). Solution of FBR model computed from exact dynamics and with learned dynamics, b). Coefficient $\mu(t)$ vs state $S(t)$, c). Coefficient $\mu(t)$ vs time $t$.}  \label{contpinns_mu_diff_ics_solve}
\end{figure}

Hence for different initial conditions and time duration beyond that of the training data, we can  solve the governing equations \eqref{eq:fedbatch} with $\mu^{NN}$ as the constitutive relation process. This idea is further illustrated in Fig. \ref{contpinns_mu_diff_ics_solve} in which the training solution data (shown in solid red) is obtained with initial conditions $[0.1, 1.0, 10.0]$ for a shorter time duration (25 sec) whereas the test solution (shown in dashed blue) is obtained with a different set of initial conditions $[0.15, 1.2, 12.0]$ for a longer duration (50 sec). Fig. \ref{fig:contpinns_mu_state_diff_ics_solve} shows the solution (in dashed black) obtained by solving governing equations \eqref{eq:fedbatch} with $\mu^{NN}(y)$ as the constitutive relation process. We can see that this solution (in dashed black) agrees very well with the test solution (in dashed blue). 

From these numerical tests, we infer that the continuous physics informed neural network (CPINN) method work well for estimating a constitutive relation but not for learning unknown dynamics. Unlike the continuous PINNS, the discrete multistep neural network (multistepNN) method works well for learning both constitutive relations and unknown dynamics. 

\section{Conclusions}
\label{sec:conclusions}
We have used both discrete (multistepNN) and continuous (CPINN) physics-informed neural network methods for learning unknown dynamics or constitutive relation of a system of ODEs. We have applied this framework to the case of a Fedbatch bioreactor. The processes in this bioreactor are very complex due to continuously changing biological and chemical reaction kinetics and nonlinear dependence of the constitutive relation on the system states. Our numerical results suggest that the discrete multistepNN framework is effective in training accurate NN-based estimators for the unknown system dynamics or constitutive relations whereas the continuous PINN framework is effective in training accurate NN-based estimators for uknown paramters alone. Our results corroborate the well-known lesson from scientific computing that whatever information we may have about the functional form of a system should be used. In particular, if we know the dynamics of a system up to an constitutive relation we can obtain more accurate predictions if we train a neural network to represent {\it only} the constitutive relation instead of the full dynamics. 

The novelty of this work is 
\begin{itemize}
    \item comparing applicability and formulation of different physics-informed machine learning approaches for learning unknown dynamics and constitutive relations of the dynamical system. 
    \item formulating and training the neural network models such that they accurately simulate the dynamics with different initial conditions and for different time duration than those of the training data. 
    \item numerical tests for highly nonlinear dynamical system (fedbatch bioreactor model) in which the constitutive relation ($\mu(S(t))$ depends on the system state $S(t)$ in a nonlinear manner. 
\end{itemize}
In future work we will study systems with larger state space and investigate the performance of the methods in the presence of noise. 

\section*{Acknowledgments}
The material presented here is based upon work supported by the Pacific Northwest National Laboratory (PNNL) ``Deep Learning for Scientific Discovery Investment". PNNL is operated by Battelle for the DOE under Contract DE-AC05-76RL01830. P.P. acknowledges support from the US Department of Energy under the Advanced Scientific Computing Research program (grant DE-SC0019116).

\bibliographystyle{elsarticle-num} 

\bibliography{references}

\end{document}